\documentclass[11pt]{article}

\usepackage[a4paper,margin=1in]{geometry}
\usepackage{palatino}
\usepackage[dvipsnames,table]{xcolor}
\usepackage[english]{babel}
\usepackage{authblk}
\usepackage{csquotes}
\usepackage{booktabs,array}
\usepackage{amsmath,amssymb}
\usepackage{graphicx}
\graphicspath{{figures/}}
\usepackage{float,placeins,caption}
\usepackage[
  backend=biber,
  style=authoryear,
  doi=false,
  url=false,
  eprint=false
]{biblatex}
\usepackage{xurl}
\usepackage[
  colorlinks=true,
  linkcolor=NavyBlue,
  citecolor=NavyBlue,
  urlcolor=NavyBlue
]{hyperref}

\definecolor{lkatblue}{HTML}{3787C0}

\AtBeginBibliography{\raggedright}
\floatstyle{ruled}
\newfloat{algorithm}{tbp}{loa}
\floatname{algorithm}{Algorithm}
\newcommand{\recipeA}{\ensuremath{\mathcal{R}_{\mathrm{A}}}}
\newcommand{\recipeB}{\ensuremath{\mathcal{R}_{\mathrm{B}}}}

\title{\bfseries
	Rethinking Vision Architectures with Gated Linear Attention and KAN
}
\date{}

\begin{document}
	
	\author[*]{Ali MEHIZEL}
	\author[$\ddagger$]{Oussama KHALDI}
	
	\affil[*]{%
		Department of Applied Statistics,
		ENSSEA, Kolea, Tipaza, Algeria\\
		\texttt{std-mehizel.ali@enssea.edu.dz}
	}
	
	\affil[$\ddagger$]{%
		Department of Computer Science,
		Université Paris Cité, Paris, France\\
		\texttt{oussama.khaldi@etu.u-paris.fr}
	}
	
	\maketitle

	\begin{abstract}
		
		Vision Transformers pour most of their parameters into MLPs for
		channel mixing, but still rely on quadratic multi-head self-attention
		for token interactions, and while linear attention fixes the
		complexity problem, bringing it down to $\mathcal{O}(N)$, it is usually
		just paired with the same fixed-activation MLP as before.
		Kolmogorov--Arnold Networks take a different approach, placing
		learnable univariate functions on the edges instead, but existing
		vision KANs either keep standard attention around or drop attention
		entirely, so nobody's really combined the two ideas well. We introduce
		\emph{LKAT} (Linear Kolmogorov--Arnold Transformer) to close that gap:
		an isotropic ViT-style encoder that couples chunkwise Gated Linear
		Attention with a two-layer KAN feed-forward block, backed by an
		I/O-aware fused RBF--KAN kernel to make the radial-basis grid functions
		actually fast in practice. Under a shared DeiT-style training recipe,
		LKAT-B outperforms ViT-B/16, ViT-5-B, and Mixer-B/16 on ImageNet-100,
		and Tiny/Small/Base variants scale consistently on CIFAR-10/100, with
		ImageNet-100 pretraining transferring cleanly to CIFAR fine-tuning,
		suggesting
		gated linear attention and KAN-based radial basis functions are
		genuinely complementary inductive biases for mid-scale visual
		representation learning.
		Code:
		\url{https://github.com/mehizelali/linear-kan-transformer}.
		
	\end{abstract}

	\noindent\textbf{Keywords:}
	Kolmogorov--Arnold Networks,
	Gated Linear Attention,
	RBF,
	Vision Architecture,
	Representation Learning

	\section{Introduction}
	
	Visual representation learning long relied on convolutional residual
	networks
	\parencite{he2016deep}; modernized ConvNets such as ConvNeXt
	\parencite{liu2022convnet} show that carefully redesigned convolutions
	remain competitive with Transformers on ImageNet and dense prediction.
	Vision Transformers instead treat images as patch sequences under an
	isotropic encoder
	\parencite{dosovitskiy2021image,touvron2021training}, building on the
	multi-head self-attention (MHSA) Transformer
	\parencite{vaswani2017attention}.
	Subsequent supervised recipes such as DeiT~III
	\parencite{touvron2022deit3} strengthen plain ViT training, while
	ViT-5 \parencite{wang2026vit5} modernizes the canonical
	Attention--FFN stack (normalization, activations, positional encoding,
	gating, and learnable tokens) without abandoning the flat topology.
	Register tokens further mitigate high-norm artifact tokens in ViT
	feature maps \parencite{darcet2024registers}. Hierarchical variants
	such as Swin and PVT
	\parencite{liu2021swin,wang2021pyramid} reintroduce multi-scale
	structure for dense tasks, while MLP-Mixer and gMLP
	\parencite{tolstikhin2021mlpmixer,liu2021gmlp} replace softmax
	self-attention with static MLP operators over tokens and channels.
	We retain a flat ViT-style topology and instead redesign the two block
	operators: attention and channel mixing.
	
	\paragraph{Linear attention.}
	Linear attention replaces the softmax similarity with a kernel feature
	map, yielding $\mathcal{O}(N)$ sequence cost and a recurrent form
	\parencite{katharopoulos2020transformers}. This recurrent view is
	formally equivalent to classical fast weight programmers, where
	additive outer products of keys and values write into a finite matrix
	memory; delta-rule updates further allow the mapping to be corrected
	over time \parencite{schlag2021linear}. Chunkwise gated mechanisms such
	as RetNet and Gated Linear Attention (GLA)
	\parencite{sun2023retnet,yang2024gated} combine inter-chunk recurrence
	with intra-chunk parallelism for hardware-efficient training. Gated Linear Attention in
	particular adds a data-dependent forget gate on the recurrent state,
	improving expressivity over plain linear attention while remaining
	subquadratic. Orthogonal to approximate or linear attention, I/O-aware
	exact attention kernels such as FlashAttention
	\parencite{dao2022flashattention} show that memory traffic
	governs wall-clock cost. These methods typically keep the
	standard two-layer MLP for channel mixing.
	
	\paragraph{Kolmogorov--Arnold Networks.}
	Kolmogorov--Arnold Networks (KANs) \parencite{liu2025kan} place
	learnable univariate functions on edges rather than fixed activations
	on nodes, motivated by the Kolmogorov--Arnold representation theorem.
	Spline-parameterized KANs are GPU-heavy; FastKAN
	\parencite{li2024fastkan} approximates edge functions with Gaussian RBFs on
	a fixed grid, enabling fused evaluation. Alternative bases (Chebyshev,
	matrix B-splines) exist
	\parencite{ss2024chebyshev,coffman2025matrixkan}, but we adopt RBF
	functions for fused GPU evaluation.
	
	\paragraph{KAN in vision.}
	Convolutional KANs \parencite{bodner2024convolutional} replace fixed CNN
	kernels with learnable spline-based univariate functions, improving parameter
	efficiency on Fashion-MNIST, but remain outside the Transformer block
	and do not address attention.
	Within ViT-style encoders, prior KAN stacks leave a clear gap relative
	to Gated Linear Attention.
	Hyb-KAN ViT \parencite{dey2025hybkan} swaps the ViT MLP for Wavelet-KAN
	(and Efficient-KAN in the head) but retains softmax MHSA; its wavelet
	path is less GPU-native than fused GEMMs.
	KAT \parencite{yang2024kat} replaces MLPs with Group-Rational KAN
	(GR-KAN) and CUDA rational kernels, yet keeps MHSA and uses a
	\emph{hybrid} KAN--MLP construction (shared group bases plus a dense
	aggregate), not a pure edge-wise KAN substitute.
	FlashKAT \parencite{raffel2025flashkat} shows that FLOPs parity with
	MLPs does not imply wall-clock parity: KAT can remain orders of
	magnitude slower in training due to memory stalls in the GR-KAN
	backward pass, motivating I/O-aware fused kernels.
	Vision KAN (ViK) \parencite{yang2026visionkan} instead removes pairwise
	attention and uses MultiPatch-RBFKAN as an attention-free sequence
	mixer; KAN-based radial-basis functions under \emph{linear} attention are left
	unexplored.
	
	Taken together, prior art either (i)~injects KANs into CNN convolutions
	rather than Transformer blocks \parencite{bodner2024convolutional},
	(ii)~pairs KAN/Wav-KAN channel mixing with quadratic softmax attention
	\parencite{dey2025hybkan,yang2024kat},
	(iii)~CUDA-accelerates a hybrid GR-KAN feed-forward while retaining MHSA,
	yet still exposing severe memory bottlenecks without careful kernels
	\parencite{raffel2025flashkat},
	(iv)~uses KAN-based radial-basis functions to replace attention rather than
	to rethink the feed-forward path under linear attention
	\parencite{yang2026visionkan}, or
	(v)~removes attention via static MLP sequence operators
	\parencite{tolstikhin2021mlpmixer,liu2021gmlp} without learnable
	univariate edge functions. The joint substitution of gated linear
	attention for MHSA and KAN-based radial-basis functions for the MLP (two complementary
	redesigns of the Transformer block) has remained largely unexamined.
	
	\paragraph{Our approach.}
	We propose \emph{LKAT} (Linear Kolmogorov--Arnold Transformer): an
	isotropic ViT backbone that couples chunkwise Gated Linear Attention with
	KAN-based radial-basis functions. We build the attention path on the
	Flash Linear Attention ({FLA}) library
	\parencite{yang2024fla}. Register tokens and an end-of-sequence class
	token keep the layout compatible with causal recurrence; absolute
	positions are added to all tokens, and 2D rotary position embeddings
	(RoPE) \parencite{su2021roformer} are applied to patch Q/K after
	per-head RMSNorm \parencite{zhang2019rmsnorm}. Motivated by
	FlashAttention's I/O-aware design
	\parencite{dao2022flashattention} and FlashKAT's analysis of KAN
	memory stalls, radial-basis grid functions are realized as I/O-aware fused
	Triton kernels \parencite{tillet2019triton} rather than unfused operators.
	We evaluate Tiny, Small, Base, and Large variants against ViT, ViT-5, and
	MLP-Mixer under a shared recipe.
	ImageNet-100 is a standard mid-scale setting for studying ViT-family
	models from scratch and for ablations
	\parencite{liu2021efficient,zheng2022vtp}; we adopt it to isolate the
	Gated Linear Attention and RBF--KAN substitutions before scaling further.
	
	Our contributions are threefold:
	\begin{itemize}
		\item We introduce LKAT, a ViT architecture that replaces multi-head
		self-attention with gated linear attention and the MLP with a
		two-layer Kolmogorov--Arnold network.
		\item We provide an I/O-aware Kolmogorov--Arnold network using Triton
		fused kernels for the radial-basis functions.
		\item We evaluate the Linear Kolmogorov--Arnold Transformer on
		mid-scale classification against ViT, ViT-5, and MLP-Mixer
		(CIFAR-10/100 and ImageNet-100), and report CIFAR fine-tuning from
		ImageNet-100 checkpoints.
	\end{itemize}

	\section{Method}
	
	\subsection{Preliminaries}
	
	We review the attention operators and Kolmogorov--Arnold constructions
	that LKAT substitutes into the ViT block.
	
	\paragraph{Multi-head self-attention (MHSA).}
	Given tokens $X\in\mathbb{R}^{N\times d}$, a single attention head
	computes queries, keys, and values
	$Q\!=\!XW_Q$, $K\!=\!XW_K$, $V\!=\!XW_V$ and
	\begin{equation}
		\mathcal{A}_{\mathrm{softmax}}(Q,K,V)
		=
		\mathrm{softmax}\!\left(\frac{QK^{\top}}{\sqrt{d_h}}\right)V,
		\label{eq:mhsa}
	\end{equation}
	with head dimension $d_h$ \parencite{vaswani2017attention}. Multi-head
	self-attention concatenates $H$ heads and applies an output projection.
	The $QK^{\top}$ term costs $\mathcal{O}(N^{2}d_h)$, which is the
	quadratic bottleneck of ViT-style encoders.
	
	\paragraph{Linear attention.}
	Linear attention removes the softmax and factorizes the similarity
	with a feature map $\phi$ \parencite{katharopoulos2020transformers}:
	\begin{equation}
		\mathcal{A}(Q,K,V)
		=
		\frac{\phi(Q)\,\big(\phi(K)^{\top}V\big)}
		{\phi(Q)\,\phi(K)^{\top}\mathbf{1}}.
		\label{eq:linattn}
	\end{equation}
	Associativity yields an equivalent recurrent form with a matrix state
	$S_t\in\mathbb{R}^{d_h\times d_h}$:
	\begin{equation}
		S_t = S_{t-1} + \phi(k_t)v_t^{\top},
		\qquad
		o_t = S_t^{\top}\phi(q_t),
		\label{eq:linattn-rnn}
	\end{equation}
	reducing sequence cost from $\mathcal{O}(N^{2})$ to $\mathcal{O}(N)$.
	
	\paragraph{Gated Linear Attention (GLA).}
	Gated Linear Attention \parencite{yang2024gated} augments the recurrent update with a
	data-dependent forget gate $G_t\in(0,1)^{d_h\times d_h}$:
	\begin{equation}
		S_t = G_t \odot S_{t-1} + k_t v_t^{\top},
		\qquad
		o_t = S_t^{\top} q_t.
		\label{eq:gla}
	\end{equation}
	For training, Gated Linear Attention uses a \emph{chunkwise-parallel} algorithm: the length-$N$
	sequence is split into chunks of size $B$; intra-chunk computation is
	parallel, while inter-chunk state is propagated serially. This preserves
	linear complexity while remaining hardware-efficient relative to naive
	recurrence.
	
	\paragraph{Kolmogorov--Arnold representation theorem.}
	The Kolmogorov--Arnold representation theorem states that any multivariate
	continuous function $f$ defined on a bounded domain can be expressed as a
	finite composition of continuous univariate functions and addition.
	In particular, for a continuous
	$f:[0,1]^{n}\!\to\!\mathbb{R}$ there exist univariate functions
	$\phi_{q,p}$ and $\Phi_{q}$ such that
	\begin{equation}
		f(x_{1},\ldots,x_{n})
		=
		\sum_{q=1}^{2n+1}
		\Phi_{q}\!
		\left(
			\sum_{p=1}^{n}
			\phi_{q,p}(x_{p})
		\right).
		\label{eq:ka-theorem}
	\end{equation}
	Equation~\eqref{eq:ka-theorem} is the classical depth-$2$ form: an inner
	layer of univariate transforms of each coordinate, followed by outer
	univariate functions of their sums. Whereas MLPs are typically motivated by
	the universal approximation theorem (fixed nonlinearities on nodes),
	KANs \parencite{liu2025kan} take Eq.~\eqref{eq:ka-theorem} as an
	architectural prior and \emph{generalize} it to deeper compositions of
	learnable univariate functions.
	
	\paragraph{Kolmogorov--Arnold Networks (KANs).}
	A modern KAN layer implements a fully connected graph in which each edge
	carries a trainable univariate function $\phi_{q,p}$ and each node merely
	sums incoming activations (no node-wise nonlinearity). Stacking such
	layers yields
	\begin{equation}
		\mathrm{KAN}(\mathbf{x})
		=
		\big(\Phi_{L}\circ\cdots\circ\Phi_{1}\big)(\mathbf{x}),
		\label{eq:kan}
	\end{equation}
	where each $\Phi_{\ell}$ is a matrix of univariate edge functions. In the
	original construction \parencite{liu2025kan}, every $\phi$ is realized as a
	residual mix of a fixed basis and a trainable spline,
	\begin{equation}
		\phi(x)
		=
		w_b\,b(x)
		+
		w_s\,\mathrm{spline}(x),
		\label{eq:kan-residual-phi}
	\end{equation}
	typically with $b(x)=\mathrm{SiLU}(x)$ and $\mathrm{spline}$ a B-spline
	expansion; $w_s$ is initialized so the spline starts near zero while $w_b$
	follows Xavier scaling, and spline grids are updated on the fly to track
	evolving activation ranges. These choices improve optimizability and local
	control, but recursive B-spline evaluation and grid adaptation are
	poorly matched to dense GPU GEMMs.
	We therefore follow FastKAN \parencite{li2024fastkan} and replace
	B-splines with Gaussian radial bases; related alternatives include
	rational bases in KAT \parencite{yang2024kat}.
	
	\paragraph{KAN-based radial-basis functions.}
	Following FastKAN, we approximate each univariate edge function by a Gaussian
	radial basis expansion on a fixed grid $\{c_g\}_{g=1}^{G}$ with
	bandwidth $d>0$.
	For a scalar activation $x$, the fused grid-linear evaluation is
	\begin{equation}
		y
		=
		\sum_{g=1}^{G}
		w_g\,
		\exp\!\left(
			-\left(\frac{x-c_g}{d}\right)^{\!2}
		\right),
		\label{eq:rbf-grid-lin}
	\end{equation}
	i.e., a weighted sum over the radial bases
	(Algorithms~\ref{alg:rbf-fwd}--\ref{alg:rbf-bwd}). This preserves the
	Kolmogorov--Arnold edge-function viewpoint while avoiding de~Boor-style
	spline recursion. Motivated by the I/O-aware design of FlashLinearAttention
	\parencite{yang2024gated}, we implement Eq.~\eqref{eq:rbf-grid-lin} as a
	fused Triton kernel rather than an unfused torch implementation that materializes
	an intermediate $[\ldots,D,G]$ RBF tensor in HBM.
	See Algorithm~\ref{alg:rbf-fwd} for the forward pass
	(Algorithm~\ref{alg:rbf-bwd} in Appendix~\ref{sec:triton-rbf} describes
	the backward pass). Autotuning selects
	$\mathrm{BLOCK}\in\{256,\ldots,4096\}$.
	In LKAT, a two-layer KAN-based radial-basis function network replaces the MLP;
	block-level pre-norm LayerNorm keeps activations within the effective
	support of the radial grid.
	
	\begin{algorithm}[H]
		\caption{Fused RBF--grid linear forward}
		\label{alg:rbf-fwd}
		\begin{tabbing}
			\hspace{1.5em}\=\hspace{1.5em}\=\hspace{1.5em}\=\kill
			\textbf{Input:} flattened $x\in\mathbb{R}^{N}$ in HBM\\
			\> centers $c\in\mathbb{R}^{G}$, weights $w\in\mathbb{R}^{G}$ in HBM\\
			\> $\mathrm{inv}\gets 1/d$; block size $\mathrm{BLOCK}$\\
			\textbf{Output:} $y\in\mathbb{R}^{N}$ in HBM\\[0.35em]
			1:\> \textbf{for} each program $p$ \textbf{in parallel do}\\
			2:\>\> Load tile $x[i]$, $i\in\mathrm{Block}(p)$ (HBM$\to$on-chip)\\
			3:\>\> $\mathrm{acc}\gets 0$ \quad\textit{(registers)}\\
			4:\>\> \textbf{for} $g=1$ \textbf{to} $G$ \textbf{do}\\
			5:\>\>\> $t\gets (x[i]-c_g)\cdot\mathrm{inv}$\\
			6:\>\>\> $\mathrm{acc}\gets\mathrm{acc}+w_g\cdot\exp(-t^{2})$\\
			7:\>\> \textbf{end for}\\
			8:\>\> Store $y[i]\gets\mathrm{acc}$ to HBM\\
			9:\> \textbf{end for}
		\end{tabbing}
	\end{algorithm}
	
\begin{figure}[H]
		\centering
		\includegraphics[width=0.92\linewidth]{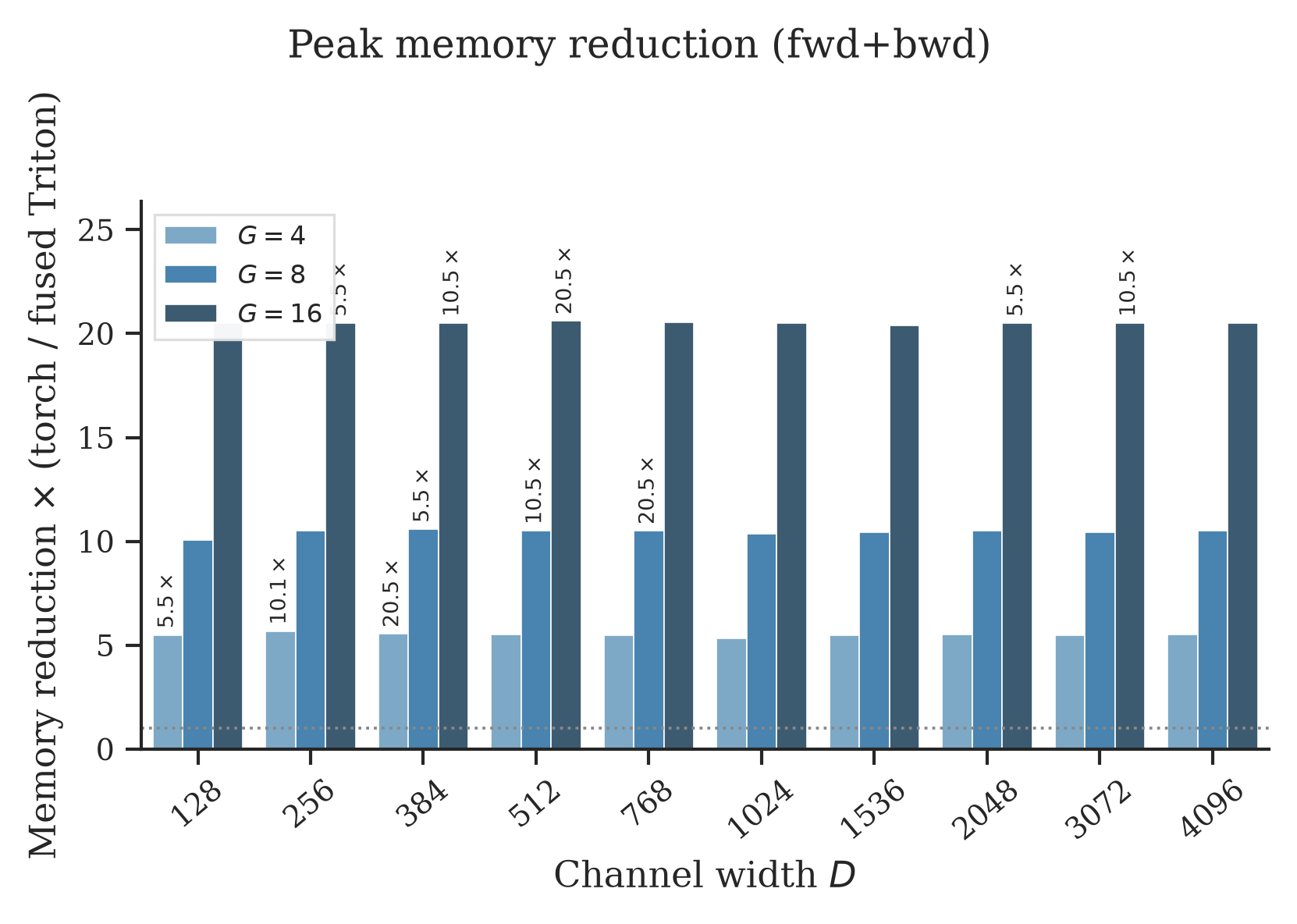}
		\caption{Peak memory reduction of the fused Triton RBF--grid linear
		operator versus an unfused torch baseline for a joint
		forward+backward pass, as a function of channel width $D$.
		Bars report $\mathrm{Mem}_{\mathrm{torch}}/\mathrm{Mem}_{\mathrm{Triton}}$
		at grid sizes $G\in\{4,8,16\}$ (dotted line ${=}1$).
		The reduction is essentially independent of $D$ and scales nearly
		linearly with $G$ ($\approx 5.5\times$, $10.5\times$, and $20.5\times$),
		consistent with avoiding materialization of the intermediate
		$[\ldots,D,G]$ expansion in HBM.}
		\label{fig:rbf-mem}
	\end{figure}

	Algorithm~\ref{alg:rbf-fwd} evaluates Eq.~\eqref{eq:rbf-grid-lin} by
	tiling the length-$N$ feature axis (batch, tokens, and channels) into
	blocks of size $\mathrm{BLOCK}$ (lines~1--2). Each program
	loads a tile of $x$ from HBM into on-chip memory and accumulates the
	weighted Gaussian bases in registers (lines~3--7), writing only the
	scalar outputs $y[i]$ back to HBM (line~8). Relative to an unfused
	composition of broadcast, elementwise exponentiation, and reduction
	(which allocates an intermediate tensor of shape $[\ldots,D,G]$ in HBM),
	this non-materializing schedule keeps the RBF expansion on chip and
	reduces HBM traffic to a single read of $x$ and a single write of $y$.
	The shared centers $c$ and weights $w$ are reused across programs; with
	$G\ll N$, the serial basis loop is inexpensive and tile-level parallelism
	maintains high GPU occupancy.
	Figure~\ref{fig:rbf-mem} reports peak memory reduction of the fused
	kernel versus an unfused torch baseline on a joint forward+backward pass:
	the reduction is essentially independent of channel width $D$ and scales
	nearly linearly with grid size $G$
	($\approx 5.5\times$ / $10.5\times$ / $20.5\times$ for $G{=}4/8/16$),
	consistent with avoiding a materialised $[\ldots,D,G]$ intermediate in HBM.
	End-to-end wall-clock speedup, which grows with both $D$ and $G$, is
	reported in Figure~\ref{fig:rbf-speedup}
	(Appendix~\ref{sec:triton-rbf}).
	
	\subsection{Linear Kolmogorov--Arnold Transformer}
	
\begin{figure}[H]
		\centering
		\includegraphics[width=0.92\linewidth,height=0.28\textheight,keepaspectratio]{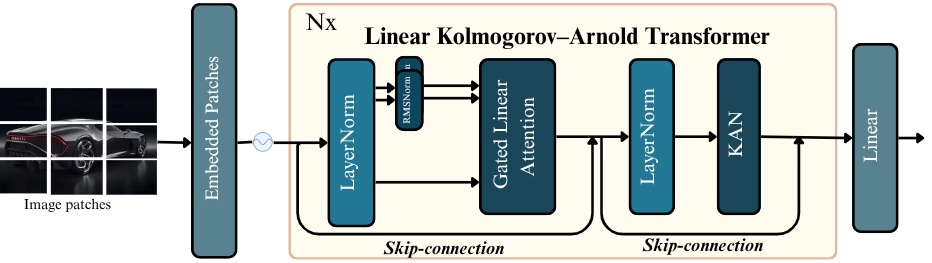}
		\caption{LKAT architecture. Patch embedding feeds $N$ pre-norm
			blocks (LayerNorm then Gated Linear Attention with per-head RMSNorm on Q/K; LayerNorm
			then KAN-based radial-basis functions), each with a residual skip.
			A linear head reads the final class token.}
		\label{fig:lkat-arch}
	\end{figure}

	We introduce LKAT (Linear Kolmogorov--Arnold Transformer), a vision
	backbone that retains the isotropic ViT topology
	\parencite{dosovitskiy2021image} while replacing multi-head
	self-attention with causal gated linear attention
	\parencite{yang2024gated,yang2024fla} and the feed-forward MLP with a two-layer
	Kolmogorov--Arnold network whose univariate bases are radial basis
	functions evaluated by I/O-aware fused Triton kernels
	(Fig.~\ref{fig:lkat-arch}; Algorithm~\ref{alg:rbf-fwd}).
	
	Given an image $\mathbf{I}\!\in\!\mathbb{R}^{H\times W\times 3}$, we
	extract non-overlapping patches of size $P\times P$ and linearly project
	them to $D$-dimensional tokens. The embedded patches are concatenated
	with a small set of learnable register tokens
	\parencite{darcet2024registers} and a class token placed
	at the sequence end, remaining compatible with causal recurrence.
	Absolute positional embeddings are added to the full sequence, while
	rotary position embeddings (2D RoPE) \parencite{su2021roformer} are
	applied only to patch query/key vectors after per-head RMS
	normalization \parencite{zhang2019rmsnorm}.
	
\begin{figure}[H]
		\centering
		\begin{minipage}[t]{0.22\linewidth}
			\centering
			\includegraphics[width=\linewidth]{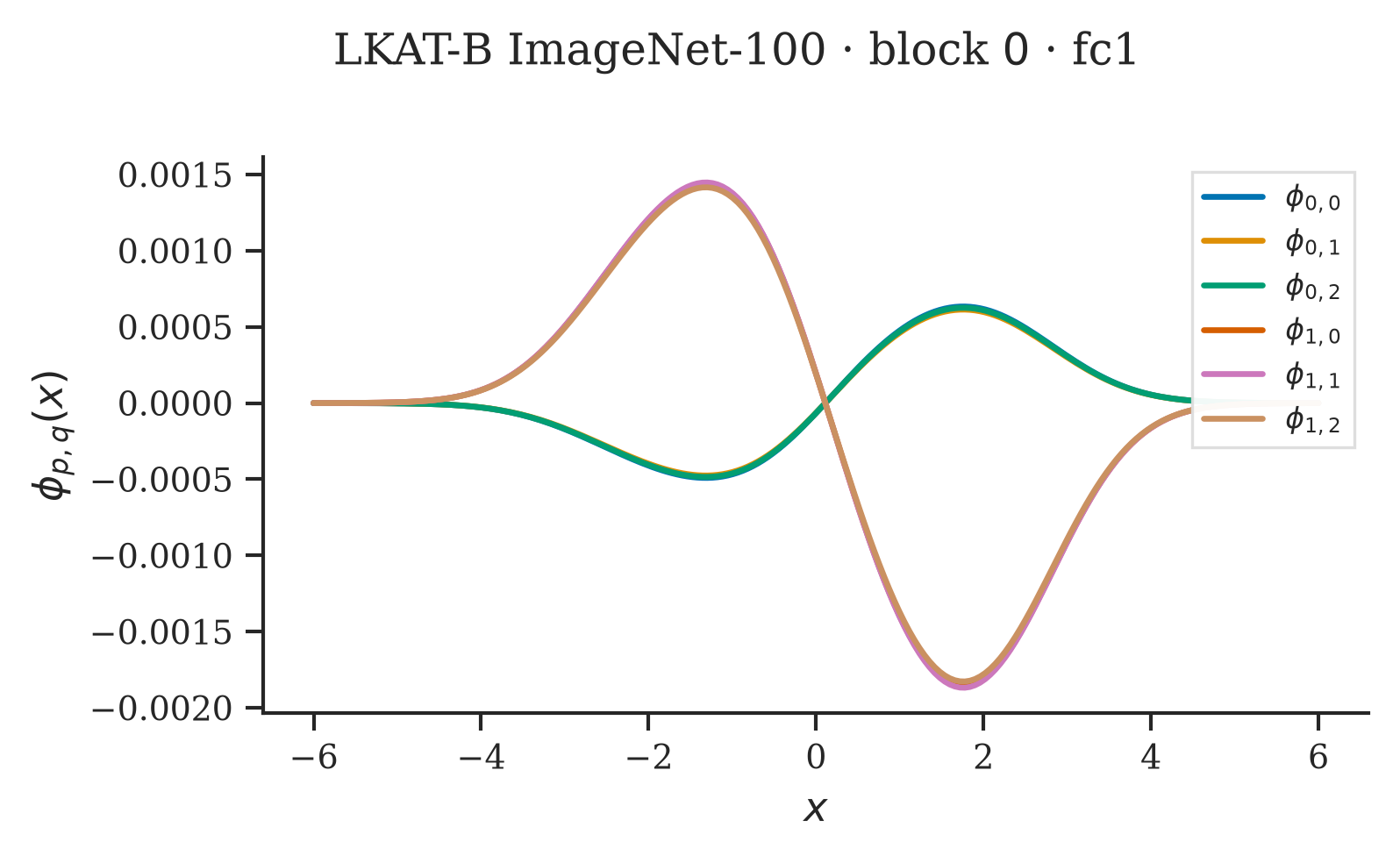}\\[-0.3em]
			{\scriptsize (a) block~$0$, fc1}
		\end{minipage}\hspace{0.03\linewidth}
		\begin{minipage}[t]{0.22\linewidth}
			\centering
			\includegraphics[width=\linewidth]{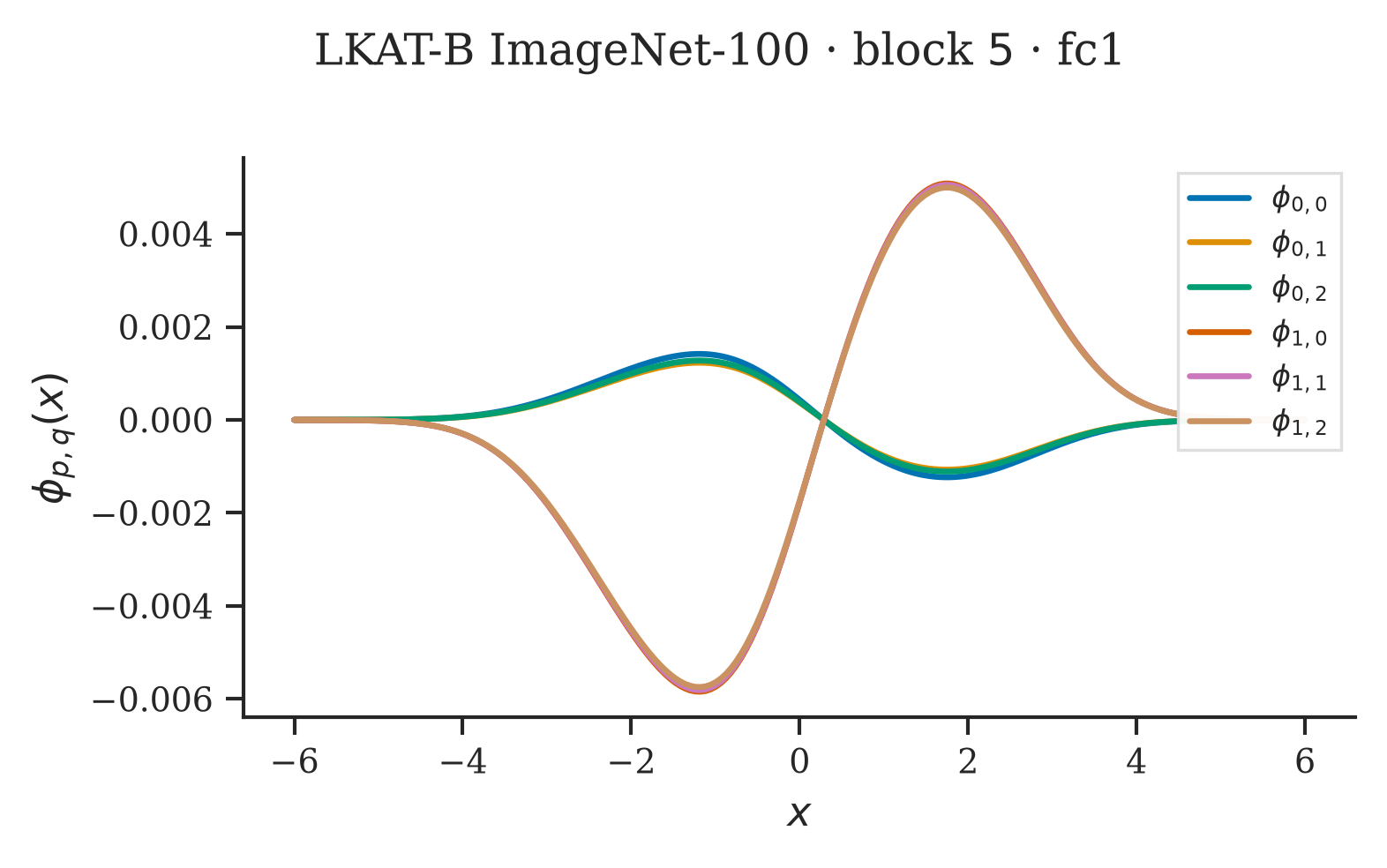}\\[-0.3em]
			{\scriptsize (b) block~$5$, fc1}
		\end{minipage}\hspace{0.03\linewidth}
		\begin{minipage}[t]{0.22\linewidth}
			\centering
			\includegraphics[width=\linewidth]{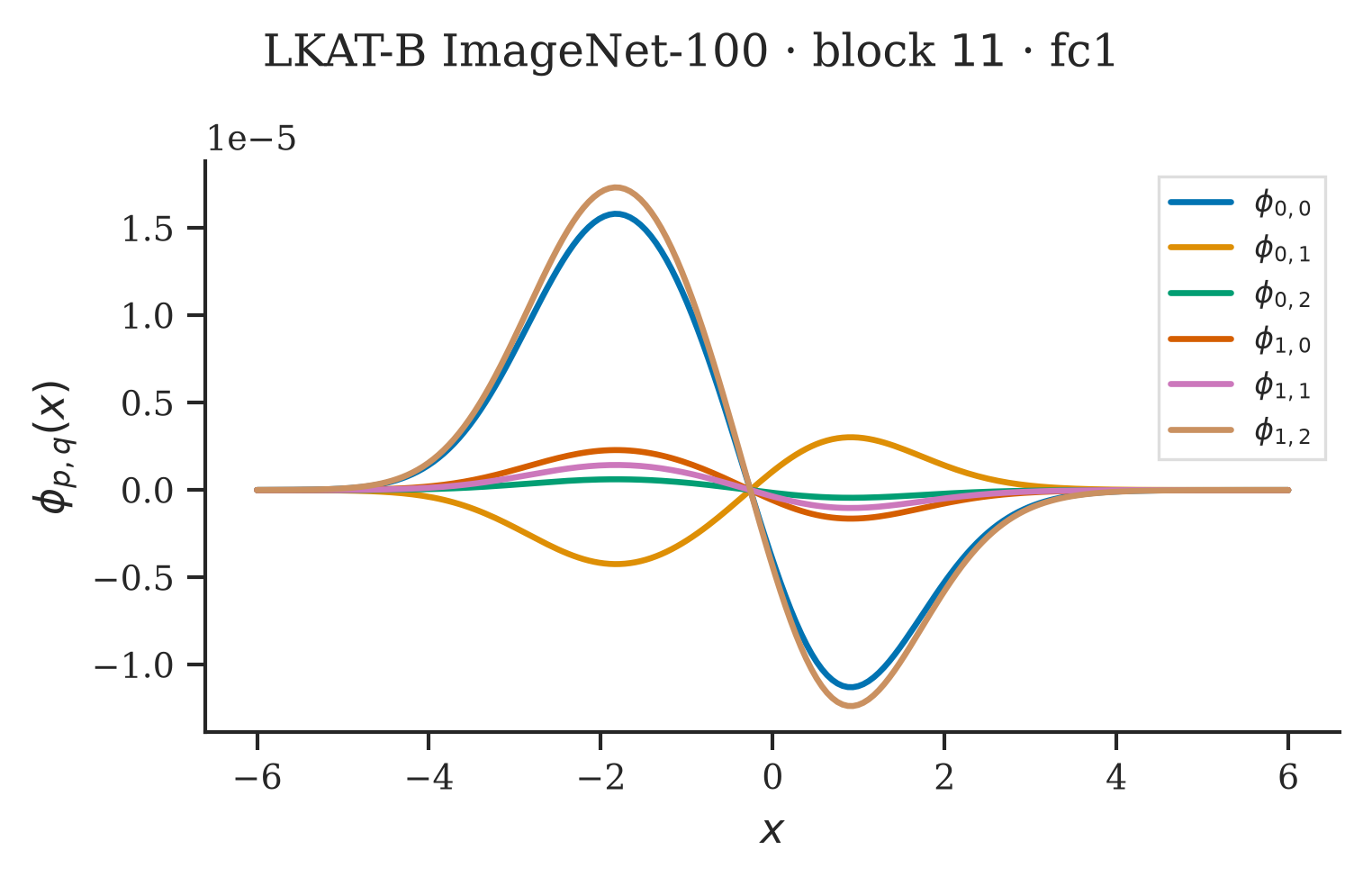}\\[-0.3em]
			{\scriptsize (c) block~$11$, fc1}
		\end{minipage}\hspace{0.03\linewidth}
		\begin{minipage}[t]{0.22\linewidth}
			\centering
			\includegraphics[width=\linewidth]{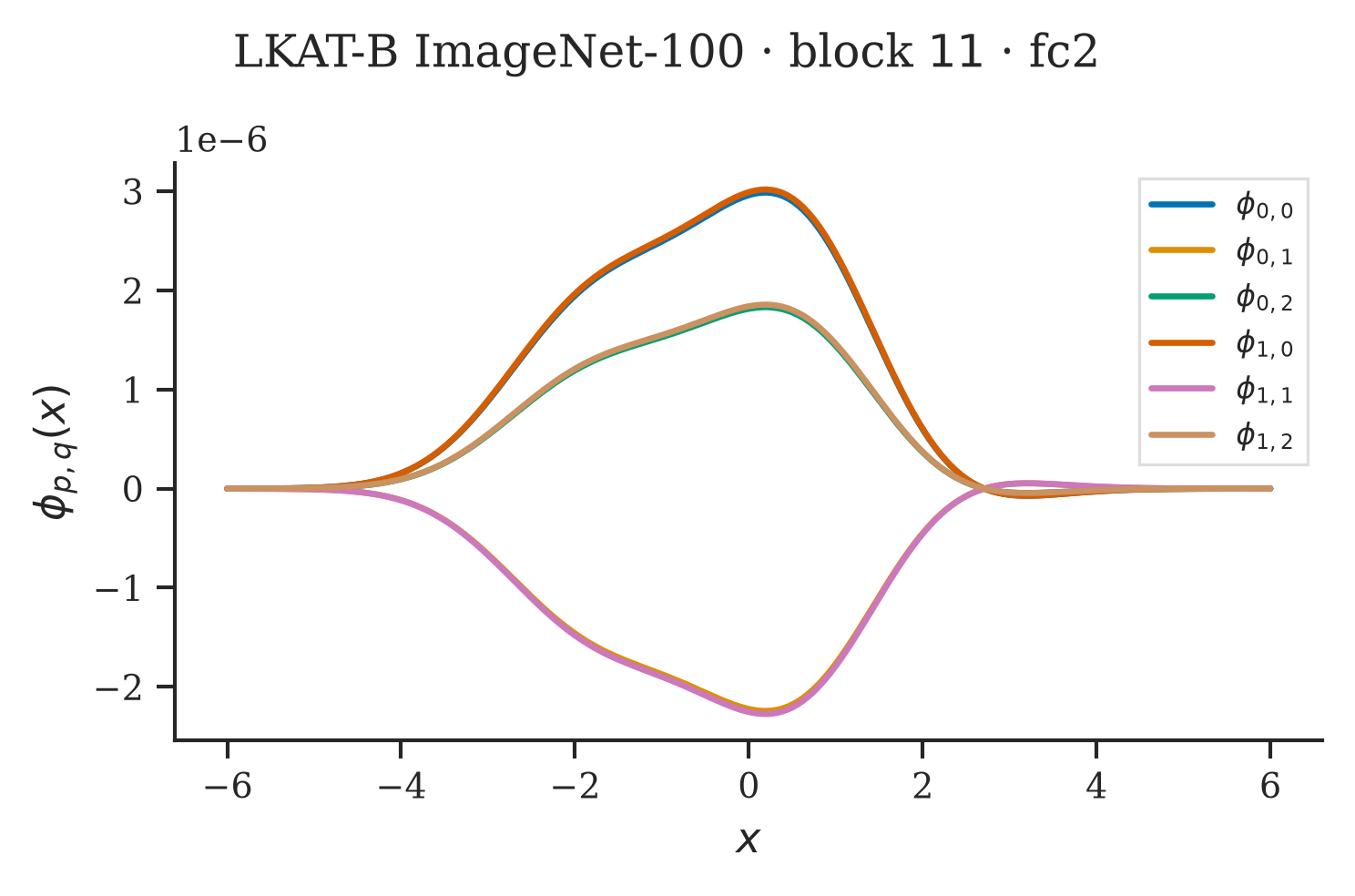}\\[-0.3em]
			{\scriptsize (d) block~$11$, fc2}
		\end{minipage}
		\caption{Learned RBF--KAN edge functions $\phi_{p,q}(x)$ for ImageNet-100
		LKAT-B (representative blocks; full set in
		Figs.~\ref{fig:learned-phi-fc1}--\ref{fig:learned-phi-fc2}).}
		\label{fig:learned-phi-main}
	\end{figure}

	Each encoder block follows a pre-normalization residual design with
	LayerNorm preceding both the Gated Linear Attention module and the KAN block
	(Fig.~\ref{fig:lkat-arch}):
	\begin{enumerate}
		\item \textbf{Gated Linear Attention:} independent linear projections
		produce queries, keys, values, and gates; short convolutions act on
		the Q/K/V paths; LayerNorm is applied on the value stream prior to
		the output projection. Attention follows Eq.~\eqref{eq:gla} in
		place of MHSA (Eq.~\eqref{eq:mhsa}).
		\item \textbf{KAN-based radial-basis functions:} a two-layer KAN with edge
		functions of the form Eq.~\eqref{eq:rbf-grid-lin}, replacing the standard
		MLP while retaining the edge-wise univariate inductive bias of
		Eqs.~\eqref{eq:ka-theorem}--\eqref{eq:kan}.
	\end{enumerate}
	Figure~\ref{fig:learned-phi-main} shows representative learned edge functions
	$\phi_{p,q}$ from ImageNet-100 LKAT-B (early, mid, and late blocks).
	The functions remain localized on the radial grid and diversify with depth;
	the full block-wise set for both KAN layers is in
	Appendix~\ref{sec:learned-curves}.
	
	Classification is performed from the final LayerNormed class token.
	Unless otherwise stated, depth, width, and patch size are matched to the
	corresponding ViT and ViT-5 baselines, yielding comparable parameter
	counts (Table~\ref{tab:results-cls}), so that accuracy differences can be
	attributed to the Gated Linear Attention and KAN-based radial-basis function substitutions rather
	than to capacity mismatch.
	
	\FloatBarrier
	
	\subsection{Experimental Setup}
	
	We train all models under one DeiT-style recipe so that accuracy gaps
	track the Gated Linear Attention and RBF--KAN substitutions rather than optimizer changes.
	Table~\ref{tab:recipe-scratch} summarizes the shared from-scratch setup.
	Following the DeiT-style pipeline
	\parencite{touvron2021training}, we use RandAugment
	\parencite{cubuk2020randaugment}, Mixup \parencite{zhang2018mixup},
	CutMix \parencite{yun2019cutmix}, Random Erasing
	\parencite{zhong2020randomerasing}, and label smoothing
	\parencite{szegedy2016rethinking}.
	
	\begin{table}[H]
		\centering
		\caption{Shared from-scratch experimental setup.
		All models (LKAT, rbfKAT, ViT, ViT-5, MLP-Mixer) use this recipe;
		only the block operators differ.}
		\label{tab:recipe-scratch}
		\renewcommand{\arraystretch}{1.15}
		\setlength{\tabcolsep}{8pt}
		\begin{tabular}{@{}ll@{}}
			\toprule
			Hyperparameter & Value \\
			\midrule
			Resolution / grid / bases & $224^{2}$ / $G{=}4$ / RBF \\
			Registers / RoPE & $4$ / 2D RoPE (\texttt{2dv1}) \\
			\midrule
			Optimizer & AdamW \\
			Peak LR / weight decay & $3{\times}10^{-4}$ / $0$ \\
			Warmup / min.\ LR & $5$ epochs / $10^{-5}$ \\
			Global batch & $256$/$512$ \\
			\midrule
			Spatial & bicubic resize $\to$ \texttt{RandomCrop}
			($P{=}4$, reflect) $\to$ HFlip \\
			RandAugment & \texttt{rand-m9-mstd0.5-inc1} \\
			Random Erasing & $p{=}0.25$, scale $(0.02,0.1)$ \\
			Mixup / CutMix & $\alpha{=}0.8$ / $1.0$ \\
			Mixup prob.\ / switch & $1.0$ / $0.5$ (batch mode) \\
			Label smoothing & $0.1$ \\
			\midrule
			Val / test pipeline & bicubic resize to
			$\texttt{input\_size}/0.875$ $\to$
			\texttt{CenterCrop} $\to$ normalize \\
			\midrule
			Epochs & CIFAR-10: $100$; CIFAR-100: $200$; ImageNet100: $300$ \\
			\bottomrule
		\end{tabular}
	\end{table}
	
	CIFAR fine-tuning from ImageNet-100 follows the appendix protocol
	(Table~\ref{tab:recipe-ft}).
	We compare \mbox{LKAT-T/S/B/L} against ViT-5, ViT, and MLP-Mixer under
	this shared setting. Full ImageNet-1K pretraining is left to future work.

	\section{Experiments}
	
	Unless otherwise stated, we train from scratch with
	Table~\ref{tab:recipe-scratch} and fine-tune with
	Table~\ref{tab:recipe-ft}, reporting best validation Top-1 accuracy (\%).
	Baselines are \textbf{ViT}, \textbf{ViT-5} \parencite{wang2026vit5}, and
	\textbf{MLP-Mixer}; \textbf{LKAT-T/S/B/L} denote our encoders with
	gated linear attention and KAN-based radial-basis functions, at parameter
	counts comparable to the matched ViT / ViT-5 scales
	(Table~\ref{tab:results-cls}). We restrict the comparison to flat ViT-style backbones and omit
	hierarchical models (e.g., Swin, PVT)
	\parencite{liu2021swin,wang2021pyramid} and modernized ConvNets
	\parencite{liu2022convnet} to isolate the effect of the block
	substitutions.
	
	\subsection{Image classification}
	
	We evaluate from-scratch Acc@1 on CIFAR-10/100 and ImageNet-100 under
	the shared recipe.
	Table~\ref{tab:results-cls} reports Top-1 Acc@1;
	Table~\ref{tab:lkat-spec} lists LKAT width, heads, and depth.
	
	\begin{table}[H]
		\centering
		\small
		\caption{From-scratch Top-1 Acc@1 (\%). Params / GFLOPs at $224^{2}$.}
		\label{tab:results-cls}
		\renewcommand{\arraystretch}{1.05}
		\setlength{\tabcolsep}{4.2pt}
		\begin{tabular}{@{}lrrccc@{}}
			\toprule
			Model & Params & GFLOPs & CIFAR-10 & CIFAR-100 & ImageNet100 \\
			\midrule
			\multicolumn{6}{@{}l}{\textit{LKAT (our)}} \\
			LKAT-T/16 & 6.25  & 2.40  & 82.24 & 62.00 & 69.74 \\
			LKAT-S/16 & 23.99 & 9.42  & 85.57 & 67.62 & 76.66 \\
			LKAT-B/16 & 93.99 & 37.34 & 88.43 & 70.25 & 80.14 \\
			\rowcolor{lkatblue!15}
			LKAT-L/16 & 329.47 & 132.14 & \textbf{90.85} & \textbf{72.10} & \textbf{83.16} \\
			\midrule
			\multicolumn{6}{@{}l}{\textit{rbfKAT (our)}} \\
			rbfKAT-T/16 & 5.51  & 2.15  & 80.96 & 61.97 & 68.92 \\
			rbfKAT-S/16 & 21.64 & 8.48  & 85.51 & 69.53 & 75.84 \\
			rbfKAT-B/16 & 85.75 & 33.70 & 89.45 & 70.41 & 79.32 \\
			\midrule
			\multicolumn{6}{@{}l}{\textit{ViT}} \\
			ViT-T/16  & 5.54  & 2.15  & 82.41 & 68.33 & 76.38 \\
			ViT-S/16  & 21.70 & 8.48  & 87.60 & 71.80 & 80.76 \\
			ViT-B/16  & 85.88 & 33.70 & 78.85 & 67.13 & 78.72 \\
			ViT-L/16 & 304.33 & 119.40 & 84.20 & 70.40 & 71.62 \\
			\midrule
			\multicolumn{6}{@{}l}{\textit{ViT-5}} \\
			ViT-5-T   & 5.54  & 2.19  & 78.46 & 58.28 & 60.26 \\
			ViT-5-S   & 21.69 & 8.65  & 82.24 & 64.84 & 72.52 \\
			ViT-5-B   & 85.85 & 34.38 & 85.70 & 68.28 & 77.46 \\
			ViT-5-L & 304.00 & 125.60 & 87.40 & 70.50 & 78.90 \\
			\midrule
			\multicolumn{6}{@{}l}{\textit{MLP-Mixer}} \\
			Mixer-T/16 & 4.16  & 1.62  & 80.80 & 65.40 & 70.80 \\
			Mixer-S/16 & 15.44 & 6.36  & 87.10 & 68.83 & 76.50 \\
			Mixer-B/16 & 59.19 & 25.20 & 88.33 & 69.86 & 78.92 \\
			Mixer-L/16 & 208.20 & 89.20 & 89.17 & 71.20 & 78.08 \\
			\bottomrule
		\end{tabular}
	\end{table}
	
	\begin{table}[H]
		\centering
		\small
		\caption{LKAT variants at $224^{2}$. Params (M) and GFLOPs for one
			forward pass.}
		\label{tab:lkat-spec}
		\renewcommand{\arraystretch}{1.05}
		\setlength{\tabcolsep}{8pt}
		\begin{tabular}{@{}lrrrrr@{}}
			\toprule
			Model & Params & GFLOPs & Dim & Heads & Depth \\
			\midrule
			LKAT-T & 6.25  & 2.40  & 192  & 3  & 12 \\
			LKAT-S & 23.99 & 9.42  & 384  & 6  & 12 \\
			LKAT-B & 93.99 & 37.34 & 768  & 12 & 12 \\
			LKAT-L & 329.47 & 132.14 & 1024 & 16 & 24 \\
			\bottomrule
		\end{tabular}
	\end{table}
	
\begin{figure}[H]
		\centering
		\includegraphics[width=0.88\linewidth]{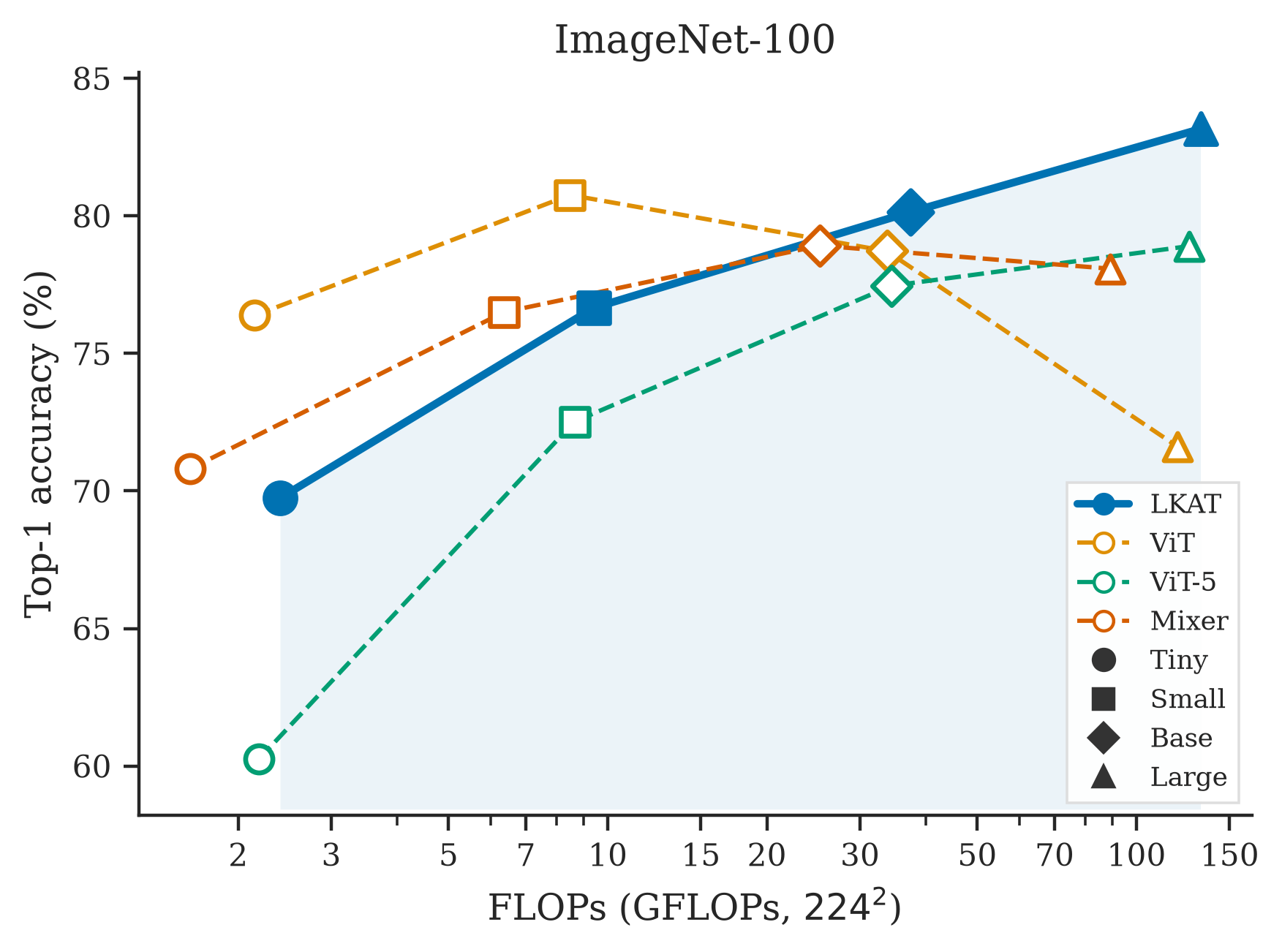}
		\caption{ImageNet-100 Top-1 Acc@1 versus GFLOPs ($224^{2}$).}
		\label{fig:scaling-in100}
	\end{figure}

	LKAT-T/S/B/L scale monotonically with capacity on CIFAR-10/100 and
	ImageNet-100. In contrast, ViT does not: under the shared recipe,
	ViT-S/16 exceeds ViT-B/16 on both CIFAR-10 and ImageNet-100, so
	increasing width/depth fails to improve Top-1 (Base underperforms
	Small). On CIFAR-10, LKAT-L leads Acc@1; among Base models, rbfKAT-B
	exceeds Mixer-L/16 and LKAT-B.
	On CIFAR-100, LKAT-L leads Acc@1; among Base models, rbfKAT-B exceeds
	LKAT-B, Mixer-B/16, and ViT-B/16. ViT-5-B surpasses ViT-B/16 at matched
	width, whereas ViT-5-T/S trail their ViT counterparts.
	On ImageNet-100, LKAT-L leads Acc@1; among Base models, LKAT-B exceeds
	rbfKAT-B, Mixer-B/16, ViT-B/16, and ViT-5-B, while ViT-S/16 exceeds
	Base and Large ViT despite fewer parameters.
	Figure~\ref{fig:scaling-in100} plots Top-1 against GFLOPs on ImageNet-100
	(CIFAR counterparts in Figs.~\ref{fig:scaling-c10}--\ref{fig:scaling-c100}).
	
	\FloatBarrier
	\section{Channel-mixing variants}
	
\begin{figure}[H]
		\centering
		\includegraphics[height=0.12\textheight,keepaspectratio]{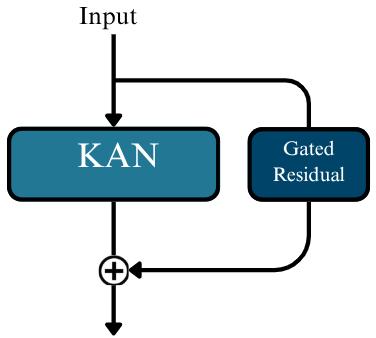}
		\caption{Gated residual around the RBF--KAN block
			(Eqs.~\eqref{eq:gated-rbf}--\eqref{eq:gated-kan}).}
		\label{fig:gated-kan}
	\end{figure}

	We ablate two controls of the RBF--KAN path on ImageNet-100 at fixed
	LKAT-B width: an additive gated residual on the RBF--KAN output, and
	the radial grid size $G$
	(Fig.~\ref{fig:gated-kan}, Table~\ref{tab:add-results}).
	The default layer applies a double weighted sum over radial bases, then
	input channels, with $\varphi_g$ as in Eq.~\eqref{eq:rbf-grid-lin}.
	The gated variant adds an elementwise residual
	$\gamma\odot\sigma(\gamma)$:
	\begin{align}
		z_p
		&=
		\sum_{g=1}^{G}
		w_{p,g}\,\varphi_g(x_p),
		\label{eq:gated-rbf}
		\\
		f
		&=
		Wz,
		\label{eq:gated-dim}
		\\
		\gamma
		&=
		W_{\gamma}x+b_{\gamma},
		\label{eq:gated-gamma}
		\\
		y
		&=
		f+\gamma\odot\sigma(\gamma).
		\label{eq:gated-kan}
	\end{align}
	
	\begin{table}[H]
		\centering
		\renewcommand{\arraystretch}{1.1}
		\setlength{\tabcolsep}{8pt}
		\begin{tabular}{@{}lccc@{}}
			\toprule
			Variant & Params (M) & GFLOPs & Acc@1 \\
			\midrule
			\multicolumn{4}{@{}l}{\emph{Gated residual}} \\
			LKAT-B & 93.99 & 37.34 & 80.14 \\
			\rowcolor{lkatblue!15}
			\quad{}+ gate & 149.96 & 60.10 & \textbf{82.02} \\
			\midrule
			\multicolumn{4}{@{}l}{\emph{RBF grid size $G$}} \\
			$G{=}4$ (default) & 93.99 & 37.34 & 80.14 \\
			$G{=}6$ & 93.29 & 37.34 & 81.08 \\
			\rowcolor{lkatblue!15}
			$G{=}8$ & 93.29 & 37.34 & \textbf{81.44} \\
			\bottomrule
		\end{tabular}
		\caption{LKAT-B ImageNet-100 ablations ($224^{2}$, 300 epochs).
			Bold denotes the best Acc@1 in each block.}
		\label{tab:add-results}
	\end{table}
	
	The gated residual improves ImageNet-100 Acc@1 over default LKAT-B
	(Table~\ref{tab:add-results}) at a larger parameter budget, indicating
	that the additive $\gamma\odot\sigma(\gamma)$ branch remains beneficial once Gated Linear Attention
	handles sequence mixing.
	Increasing the radial grid from $G{=}4$ to $G{=}6$ and $G{=}8$ likewise
	raises Acc@1 above the default LKAT-B entry in
	Table~\ref{tab:results-cls}, consistent with under-resolved univariate
	bases under the compact default grid; denser grids trade expressivity
	for additional basis parameters and fused RBF work.
	
	\section{CIFAR fine-tuning}
	
	We initialize from ImageNet-100 checkpoints and fine-tune the full
	backbone. CIFAR inputs are bilinearly upsampled to $224^{2}$ by default
	($384^{2}$ for higher-resolution \recipeB).
	Table~\ref{tab:downstream-cifar} reports CIFAR transfer from the
	ImageNet-100 LKAT-B checkpoint.
	
	\begin{table}[H]
		\centering
		\caption{CIFAR fine-tuning of ImageNet-100 LKAT-B.}
		\label{tab:downstream-cifar}
		\renewcommand{\arraystretch}{1.15}
		\setlength{\tabcolsep}{8pt}
		\begin{tabular}{@{}lcc@{}}
			\toprule
			Setting & CIFAR-10 & CIFAR-100 \\
			\midrule
			\recipeA, $224^{2}$ & 93.01 & 78.20 \\
			\recipeB, $224^{2}$ & 93.62 & 78.28 \\
			\rowcolor{lkatblue!15}
			\recipeB, $384^{2}$ & \textbf{95.45} & \textbf{78.63} \\
			\bottomrule
		\end{tabular}
	\end{table}
	
	LKAT-B fine-tuning under \recipeA/\recipeB~at $224^{2}$ improves
	substantially over from-scratch CIFAR training, and \recipeB~at
	$384^{2}$ yields a further gain on both CIFAR-10 and CIFAR-100
	(Table~\ref{tab:downstream-cifar}).
	
	\FloatBarrier

	\section{Discussion}
	
	LKAT-T/S/B/L exhibit monotonic capacity scaling on CIFAR and ImageNet-100
	(Table~\ref{tab:results-cls}), whereas ViT plateaus or declines from
	Small to Base under the same recipe (ViT-S/16 outperforms ViT-B/16 on
	CIFAR-10 and ImageNet-100), indicating a mid-scale optimization /
	capacity mismatch for plain MHSA--MLP stacks. LKAT-B remains competitive
	against ViT-B, ViT-5-B, and Mixer-B.
	Ablations show that a gated KAN residual and a denser RBF grid
	both improve ImageNet-100 Acc@1 relative to the default
	block (Table~\ref{tab:add-results}).
	Fine-tuning from ImageNet-100 yields
	substantial Acc@1 improvements over from-scratch training
	(Table~\ref{tab:downstream-cifar}), indicating that the joint
	gated-linear-attention and KAN-based radial-basis representation transfers
	within classification.
	These results motivate full ImageNet-1K pretraining under the same block
	design, following the common ablation-then-scale pattern on ImageNet-100
	\parencite{zheng2022vtp}.

	\section{Conclusion}
	
	We presented LKAT, an isotropic vision encoder that replaces MHSA with
	chunkwise Gated Linear Attention and the MLP with a two-layer
	KAN-based radial-basis function network, under a shared DeiT-style
	training recipe.
	The design isolates complementary inductive biases for attention and
	channel mixing without hierarchical stages or dense-prediction heads.
	
	Empirically, LKAT-T/S/B/L scale monotonically on CIFAR-10/100 and ImageNet-100
	(Table~\ref{tab:results-cls}), while ViT does not: ViT-S/16 exceeds
	ViT-B/16 on CIFAR-10 and ImageNet-100 under the shared recipe. On
	ImageNet-100, LKAT-B exceeds rbfKAT-B, ViT-B/16, ViT-5-B, and Mixer-B/16; LKAT-L
	leads Acc@1 on CIFAR-10/100 and ImageNet-100.
	ImageNet-100 pretraining transfers effectively: \recipeA/\recipeB
	at $224^{2}$ and \recipeB~at $384^{2}$ yield substantial Acc@1 improvements over from-scratch
	CIFAR training (Table~\ref{tab:downstream-cifar}).
	
	These results indicate that gated linear attention and KAN-based radial-basis
	functions are effective substitutes for the canonical ViT block for
	mid-scale visual representation learning, with more reliable capacity
	scaling than softmax MHSA under the same DeiT-style recipe. Future work
	includes full ImageNet-1K pretraining under the same block design
	\parencite{zheng2022vtp} and broader transfer studies.
	
	\section*{Acknowledgements}
	
	We thank Ayoub Asri and Lizhong Chen for their
	support, and AI~GRID for providing computational resources, including
	NVIDIA RTX~5090 GPUs, which enabled the from-scratch training,
	ImageNet-100 pretraining, and CIFAR fine-tuning experiments reported in
	this work.

	\newpage
	\section*{Appendix}
	\label{sec:appendix}
	
	This appendix reports the CIFAR fine-tuning recipes, data preprocessing,
	CIFAR scaling curves, fused RBF--grid kernels, and learned
	edge-function visualizations.
	
	\subsection*{Fine-tuning recipe}
	Table~\ref{tab:recipe-ft} details CIFAR fine-tuning from the ImageNet-100
	LKAT-B checkpoint under \recipeA~and \recipeB.
	
	\begin{table}[H]
		\centering
		\caption{CIFAR fine-tuning recipe ($\recipeA$ vs.\ $\recipeB$).}
		\label{tab:recipe-ft}
		\renewcommand{\arraystretch}{1.12}
		\setlength{\tabcolsep}{8pt}
		\begin{tabular}{@{}lll@{}}
			\toprule
			Setting & \recipeA & \recipeB \\
			\midrule
			\multicolumn{3}{@{}l}{\textit{Shared}} \\
			Init & \multicolumn{2}{c}{ImageNet-100 pretrained encoder} \\
			Optimizer / peak LR & \multicolumn{2}{c}{AdamW / $10^{-4}$} \\
			Resolution / aug. & \multicolumn{2}{c}{$224^{2}$/$384^{2}$ / DeiT} \\
			Mixup / CutMix & \multicolumn{2}{c}{$0.8$ / $1.0$} \\
			\midrule
			\multicolumn{3}{@{}l}{\textit{Differing}} \\
			Weight decay & $0$ & $0.05$ \\
			Drop path & $0$ & $0.1$ \\
			KAN LayerNorm & on & off \\
			\midrule
			CIFAR-10 Acc@1 & $93.01$ & $\mathbf{93.62}$ \\
			\bottomrule
		\end{tabular}
	\end{table}
	
	\subsection*{Data preprocessing}
	\label{sec:aug-pipeline}
	
	Let $S$ denote the target resolution ($S{=}224$ from scratch and for
	$224^{2}$ fine-tuning; $S{=}384$ for higher-resolution \recipeB).
	Training applies bicubic resize to $S$, random crop ($S$, padding~$4$,
	reflect), horizontal flip, RandAugment (\texttt{rand-m9-mstd0.5-inc1}),
	normalization, and random erasing ($p{=}0.25$, scale $(0.02,0.1)$),
	then batch Mixup/CutMix ($\alpha{=}0.8/1.0$, probability~$1.0$,
	switch~$0.5$, batch mode) with label smoothing~$0.1$.
	Validation/test use bicubic resize to $S/0.875$, center crop to $S$,
	and the same normalization.
	
	\newpage
	\subsection*{CIFAR scaling curves}
	
\begin{figure}[H]
		\centering
		\begin{minipage}[t]{0.485\linewidth}
			\centering
			\includegraphics[width=\linewidth]{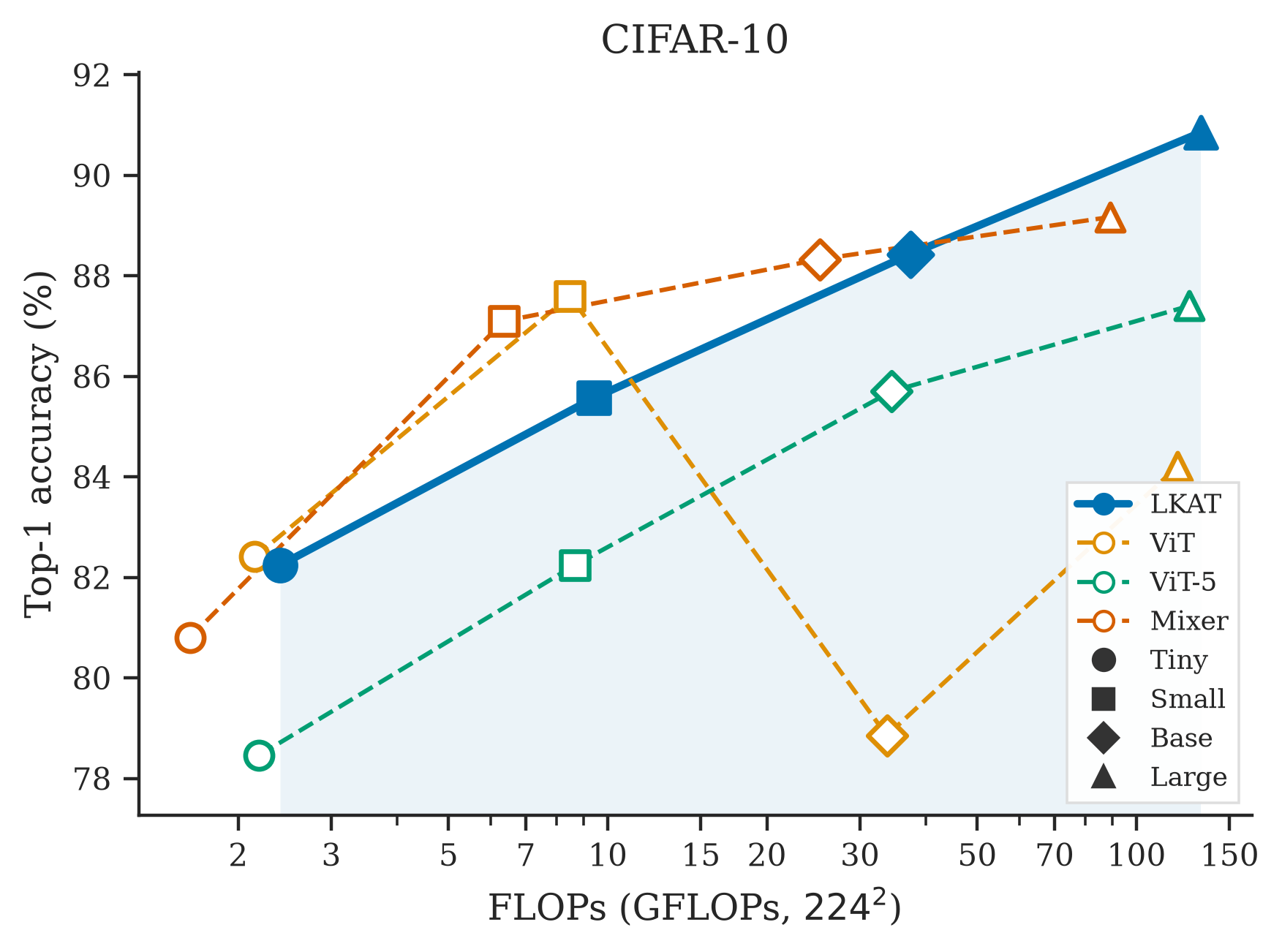}
			\caption{CIFAR-10 Top-1 Acc@1 versus GFLOPs ($224^{2}$).}
			\label{fig:scaling-c10}
		\end{minipage}\hfill
		\begin{minipage}[t]{0.485\linewidth}
			\centering
			\includegraphics[width=\linewidth]{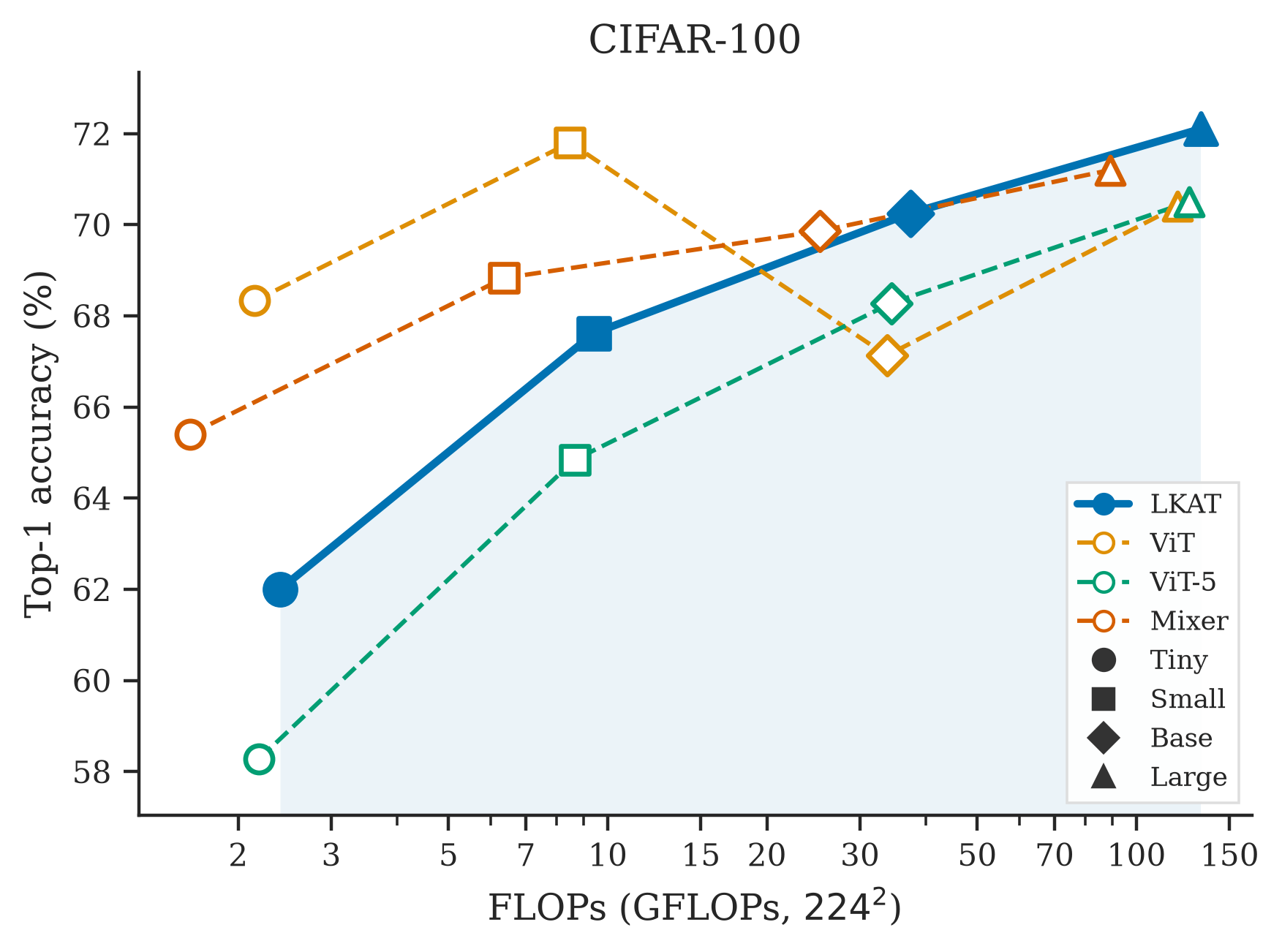}
			\caption{CIFAR-100 Top-1 Acc@1 versus GFLOPs ($224^{2}$).}
			\label{fig:scaling-c100}
		\end{minipage}
	\end{figure}

	Figures~\ref{fig:scaling-c10}--\ref{fig:scaling-c100} extend the
	ImageNet-100 scaling plot in the main text.
	
	\subsection*{Fused RBF--grid linear kernels}
	\label{sec:triton-rbf}

\begin{figure}[H]
		\centering
		\includegraphics[width=\linewidth]{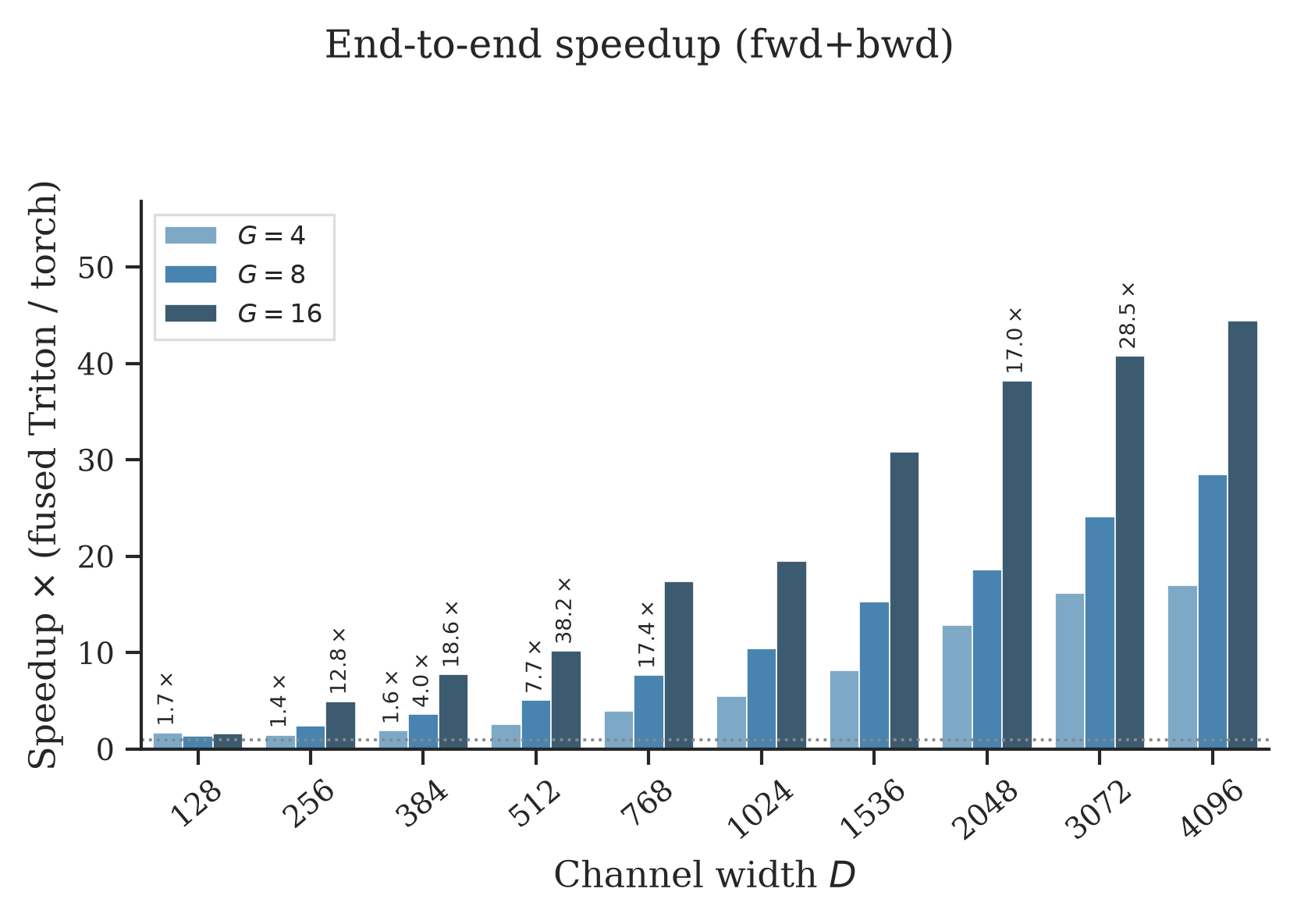}
		\caption{End-to-end wall-clock speedup of the fused Triton RBF--grid
		linear operator versus an unfused torch baseline for a joint
		forward+backward pass, as a function of channel width $D$.
		Bars report $\mathrm{Time}_{\mathrm{torch}}/\mathrm{Time}_{\mathrm{Triton}}$
		at grid sizes $G\in\{4,8,16\}$ (dotted line ${=}1$).
		Unlike peak-memory reduction, which is nearly constant in $D$,
		e2e speedup grows with both $D$ and $G$: from about
		$1.4$--$1.7\times$ at $D{=}128$ up to about
		$17\times$ / $29\times$ / $44\times$ at $D{=}4096$ for
		$G{=}4/8/16$. Larger grids amplify the benefit of avoiding the
		materialised $[\ldots,D,G]$ intermediate, so the fused kernel's
		latency stays nearly flat while torch scales with $G$.}
		\label{fig:rbf-speedup}
	\end{figure}

	The forward kernel is given in Algorithm~\ref{alg:rbf-fwd}.
	We describe the fused backward pass below and report end-to-end
	wall-clock speedup versus an unfused torch baseline
	(Figure~\ref{fig:rbf-speedup}), in the same
	spirit as the FlashLinearAttention microbenchmarks of
	\parencite{yang2024gated}.
	Peak memory reduction appears in the main text
	(Figure~\ref{fig:rbf-mem}).
	Our reference implementation uses Triton
	\parencite{tillet2019triton}.
	
	\paragraph{Backpropagation.}
	Write the scalar function of Eq.~\eqref{eq:rbf-grid-lin} as
	\begin{equation}
		y
		=
		\sum_{g=1}^{G}
		w_g\,\varphi_g(x),
		\qquad
		\varphi_g(x)
		=
		\exp\!\left(-\left(\frac{x-c_g}{d}\right)^{2}\right).
		\label{eq:rbf-fwd-compact}
	\end{equation}
	Differentiating w.r.t.\ the input and the basis weights yields
	\begin{align}
		\frac{\partial y}{\partial x}
		&=
		\sum_{g=1}^{G}
		w_g\,\varphi_g(x)\,
		\left(
			-2\,\frac{x-c_g}{d^{2}}
		\right),
		\label{eq:rbf-dx}
		\\
		\frac{\partial y}{\partial w_g}
		&=
		\varphi_g(x).
		\label{eq:rbf-dw}
	\end{align}
	Given an upstream gradient $\mathrm{d}y=\frac{\partial\mathcal{L}}{\partial y}$,
	the local gradients are
	\begin{align}
		\mathrm{d}x
		&=
		\mathrm{d}y\cdot\frac{\partial y}{\partial x},
		\label{eq:rbf-dx-loss}
		\\
		\mathrm{d}w_g
		&=
		\sum_{i=1}^{N}
		\mathrm{d}y_i\,\varphi_g(x_i).
		\label{eq:rbf-dw-loss}
	\end{align}
	The fused backward kernel (Algorithm~\ref{alg:rbf-bwd}) recomputes
	$\{\varphi_g\}$ on the fly, accumulates Eq.~\eqref{eq:rbf-dx} in
	registers for $\mathrm{d}x$, and aggregates Eq.~\eqref{eq:rbf-dw-loss}
	with atomic adds over the tiled feature axis.
	
	\begin{algorithm}[H]
		\caption{Fused RBF--grid linear backward}
		\label{alg:rbf-bwd}
		\begin{tabbing}
			\hspace{1.5em}\=\hspace{1.5em}\=\hspace{1.5em}\=\kill
			\textbf{Require:} $x,c,w,d$ as in Alg.~\ref{alg:rbf-fwd};\\
			\> upstream $\mathrm{d}y\in\mathbb{R}^{N}$\\
			\textbf{Ensure:} $\mathrm{d}x\in\mathbb{R}^{N}$,
			$\mathrm{d}w\in\mathbb{R}^{G}$\\[0.35em]
			1:\> $\mathrm{inv}\gets 1/d$;\
			$\mathrm{inv}^{2}\gets 1/d^{2}$;\
			$\mathrm{d}w\gets 0$\\
			2:\> \textbf{for} each program $p$ \textbf{in parallel do}\\
			3:\>\> Load tile $x[i]$, $\mathrm{d}y[i]$\\
			4:\>\> $\mathrm{jac}\gets 0$\\
			5:\>\> \textbf{for} $g=1$ \textbf{to} $G$ \textbf{do}\\
			6:\>\>\> $\mathrm{diff}\gets x[i]-c_g$\\
			7:\>\>\> $r\gets\exp\!\big(-(\mathrm{diff}\cdot\mathrm{inv})^{2}\big)$\\
			8:\>\>\> $\mathrm{jac}\gets\mathrm{jac}
w_g\,r\,(-2\cdot\mathrm{diff}\cdot\mathrm{inv}^{2})$\\
			9:\>\>\> $\mathrm{AtomicAdd}(\mathrm{d}w_g,\;\sum_{i}\mathrm{d}y[i]\cdot r)$\\
			10:\>\> \textbf{end for}\\
			11:\>\> Store $\mathrm{d}x[i]\gets\mathrm{d}y[i]\cdot\mathrm{jac}$\\
			12:\> \textbf{end for}
		\end{tabbing}
	\end{algorithm}
	
	\subsection*{Learned RBF--KAN edge functions}
	\label{sec:learned-curves}
	
\begin{figure}[H]
		\centering
		\begin{minipage}[t]{0.15\linewidth}
			\centering
			\includegraphics[width=\linewidth]{learned_curve/lkat_b_imagenet100/block00_fc1_curves.png}\\[-0.25em]
			{\scriptsize 0}
		\end{minipage}\hspace{0.01\linewidth}
		\begin{minipage}[t]{0.15\linewidth}
			\centering
			\includegraphics[width=\linewidth]{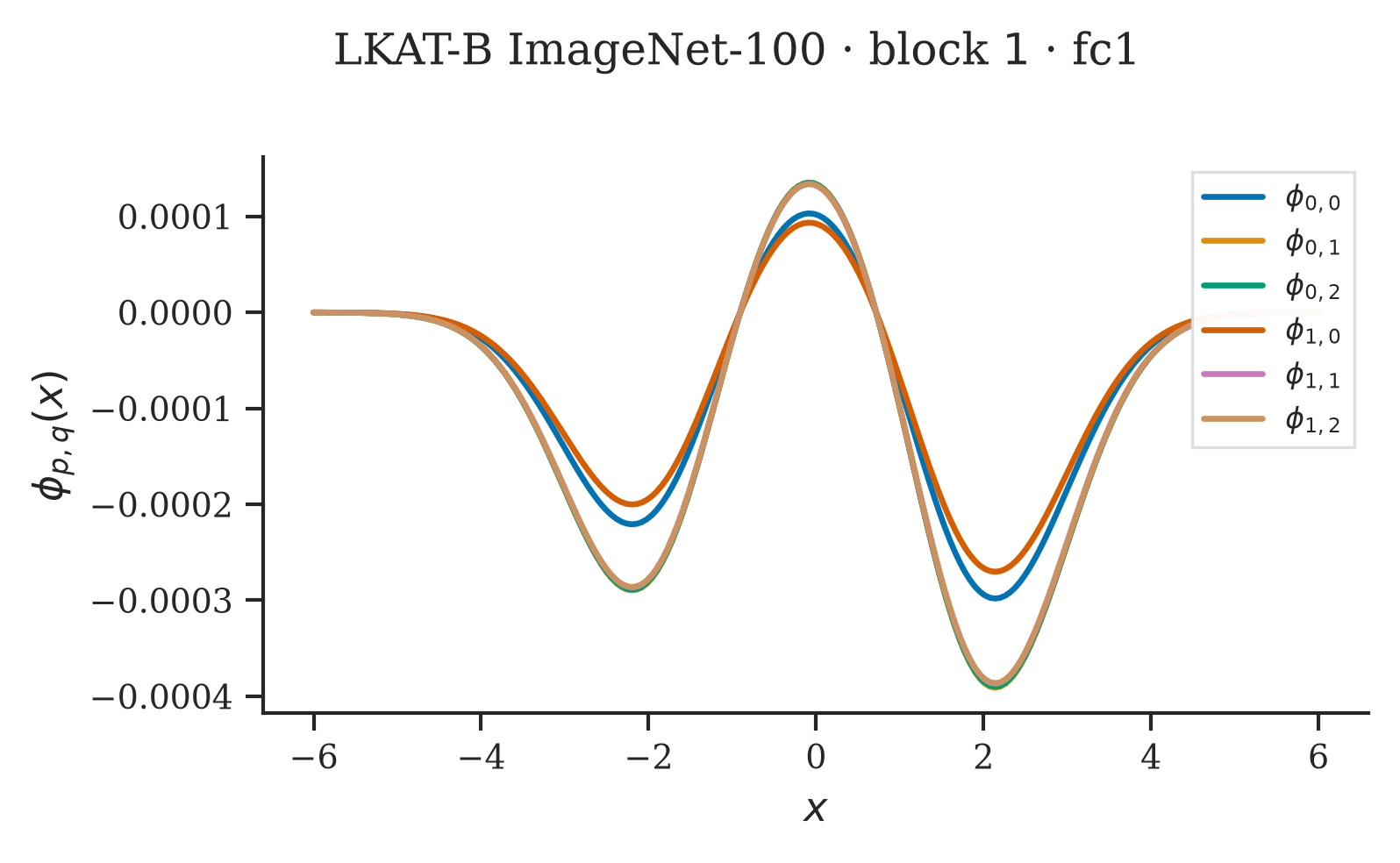}\\[-0.25em]
			{\scriptsize 1}
		\end{minipage}\hspace{0.01\linewidth}
		\begin{minipage}[t]{0.15\linewidth}
			\centering
			\includegraphics[width=\linewidth]{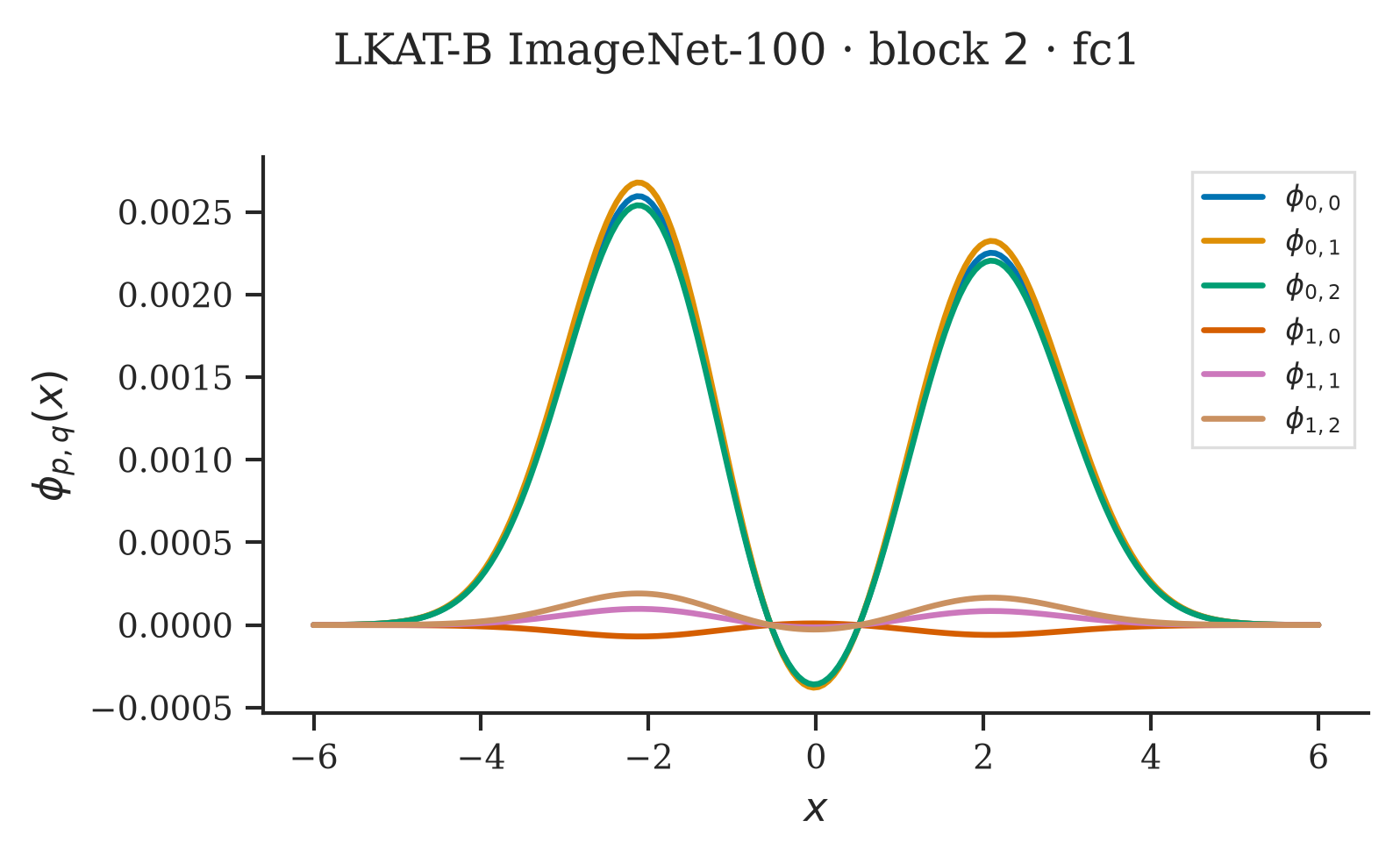}\\[-0.25em]
			{\scriptsize 2}
		\end{minipage}\hspace{0.01\linewidth}
		\begin{minipage}[t]{0.15\linewidth}
			\centering
			\includegraphics[width=\linewidth]{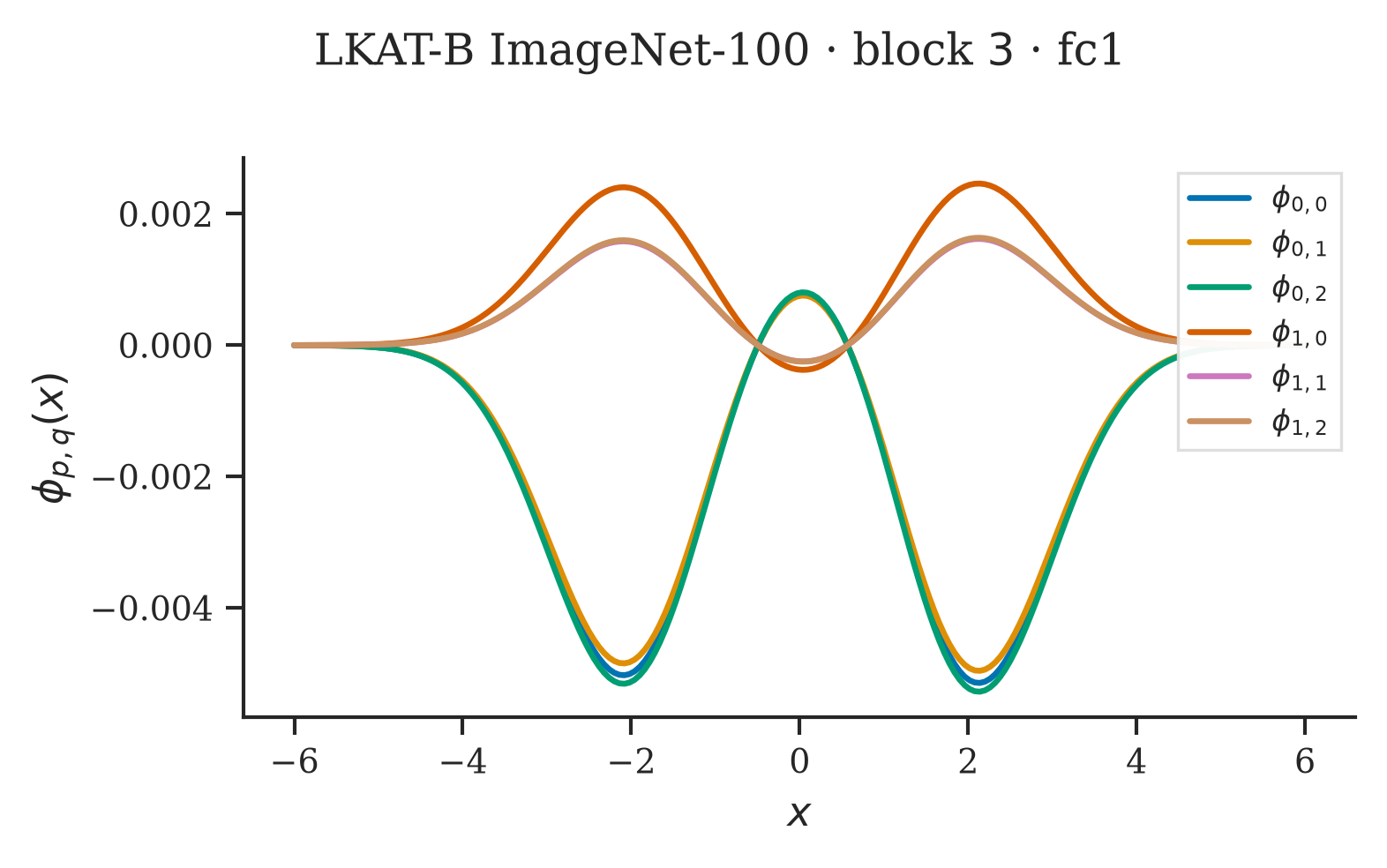}\\[-0.25em]
			{\scriptsize 3}
		\end{minipage}\hspace{0.01\linewidth}
		\begin{minipage}[t]{0.15\linewidth}
			\centering
			\includegraphics[width=\linewidth]{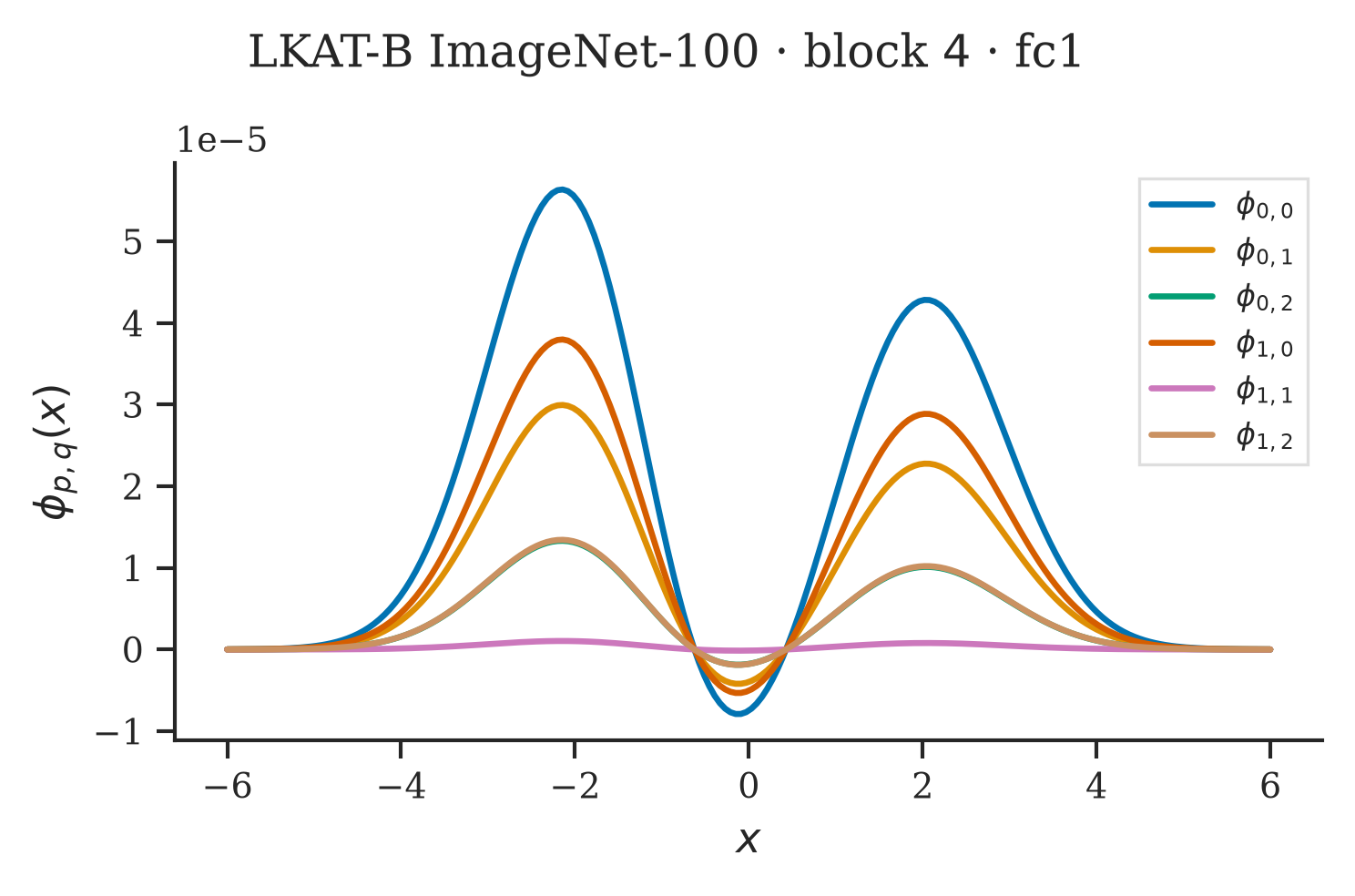}\\[-0.25em]
			{\scriptsize 4}
		\end{minipage}\hspace{0.01\linewidth}
		\begin{minipage}[t]{0.15\linewidth}
			\centering
			\includegraphics[width=\linewidth]{learned_curve/lkat_b_imagenet100/block05_fc1_curves.png}\\[-0.25em]
			{\scriptsize 5}
		\end{minipage}\\[0.35em]
		\begin{minipage}[t]{0.15\linewidth}
			\centering
			\includegraphics[width=\linewidth]{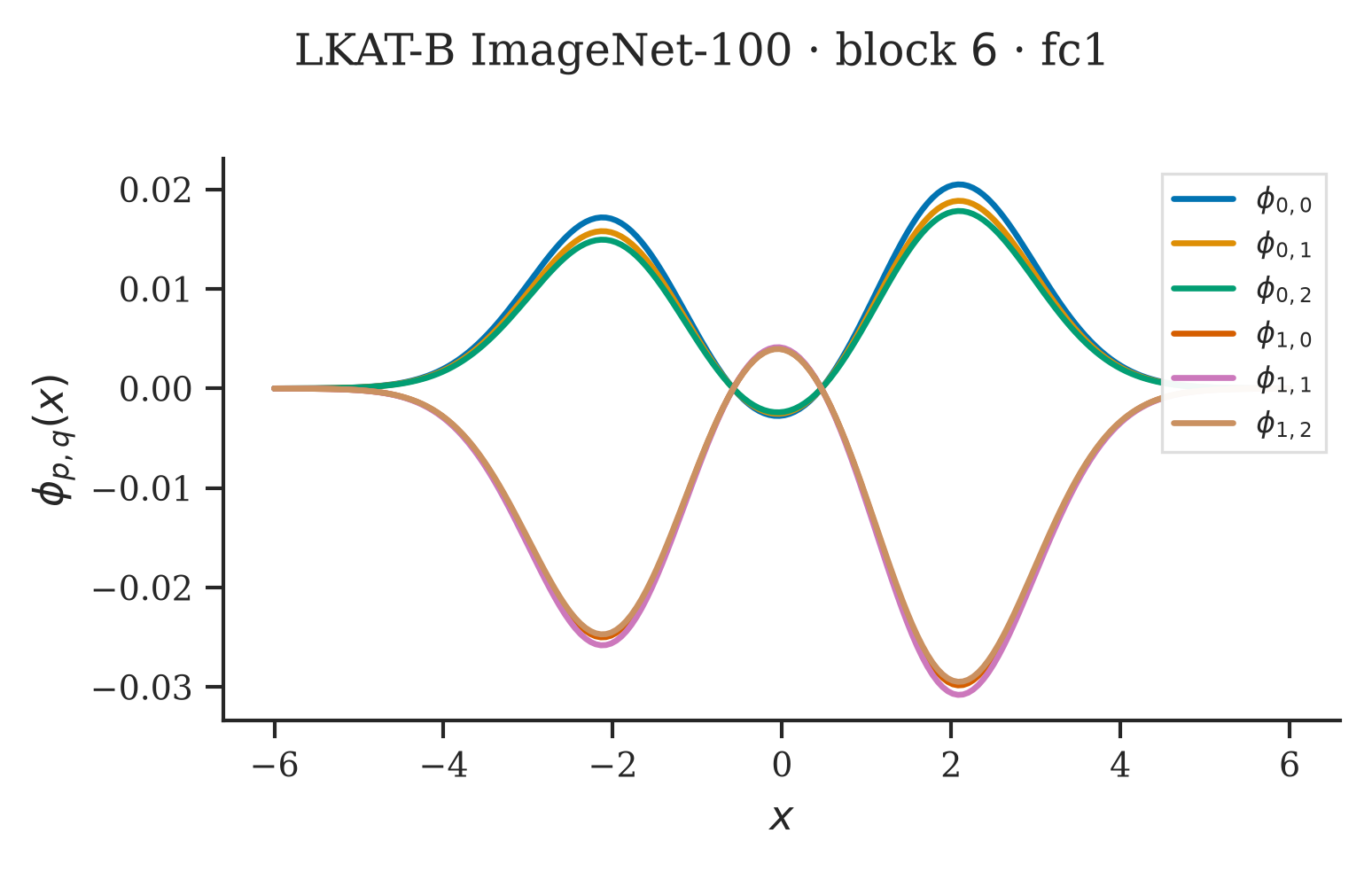}\\[-0.25em]
			{\scriptsize 6}
		\end{minipage}\hspace{0.01\linewidth}
		\begin{minipage}[t]{0.15\linewidth}
			\centering
			\includegraphics[width=\linewidth]{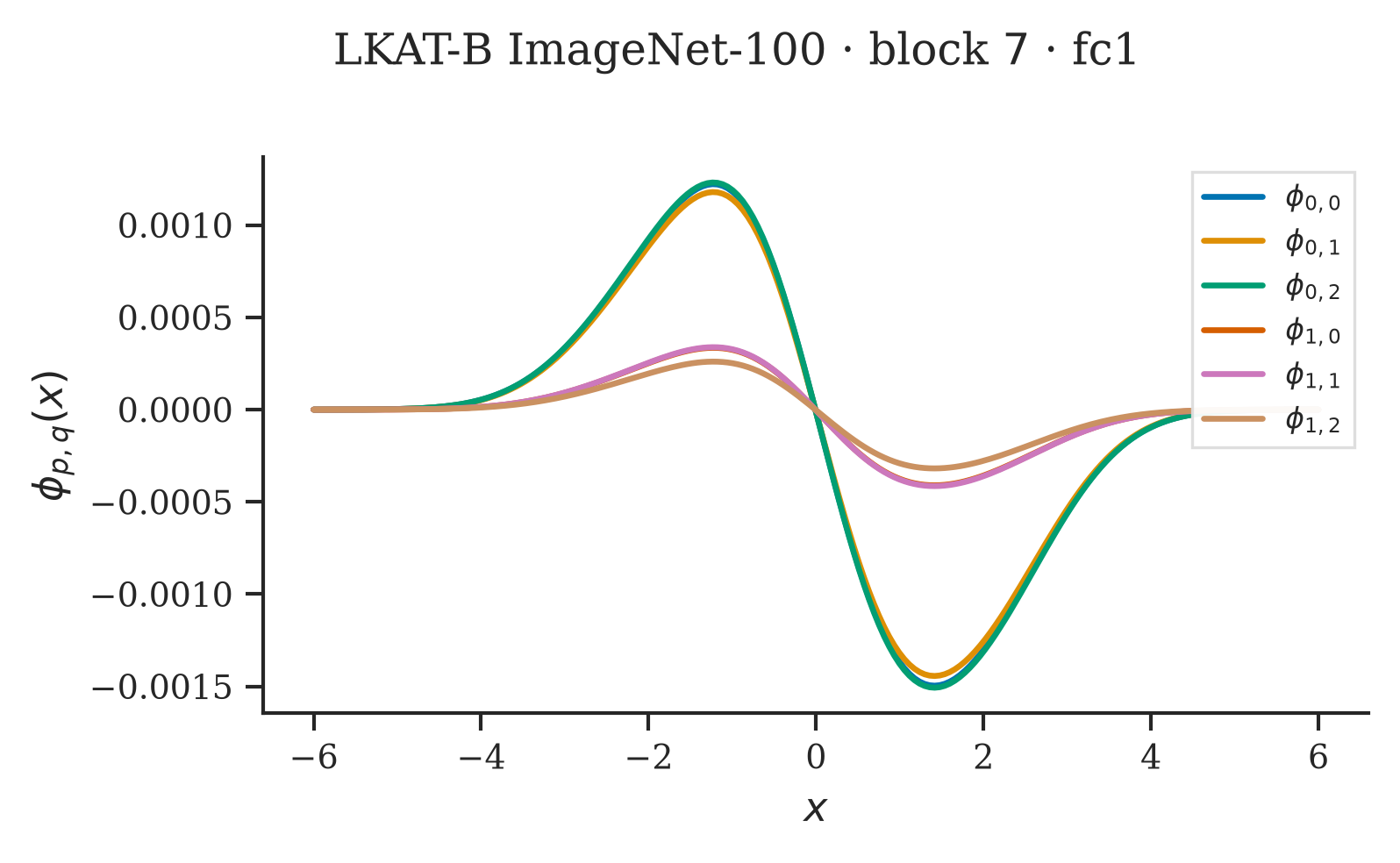}\\[-0.25em]
			{\scriptsize 7}
		\end{minipage}\hspace{0.01\linewidth}
		\begin{minipage}[t]{0.15\linewidth}
			\centering
			\includegraphics[width=\linewidth]{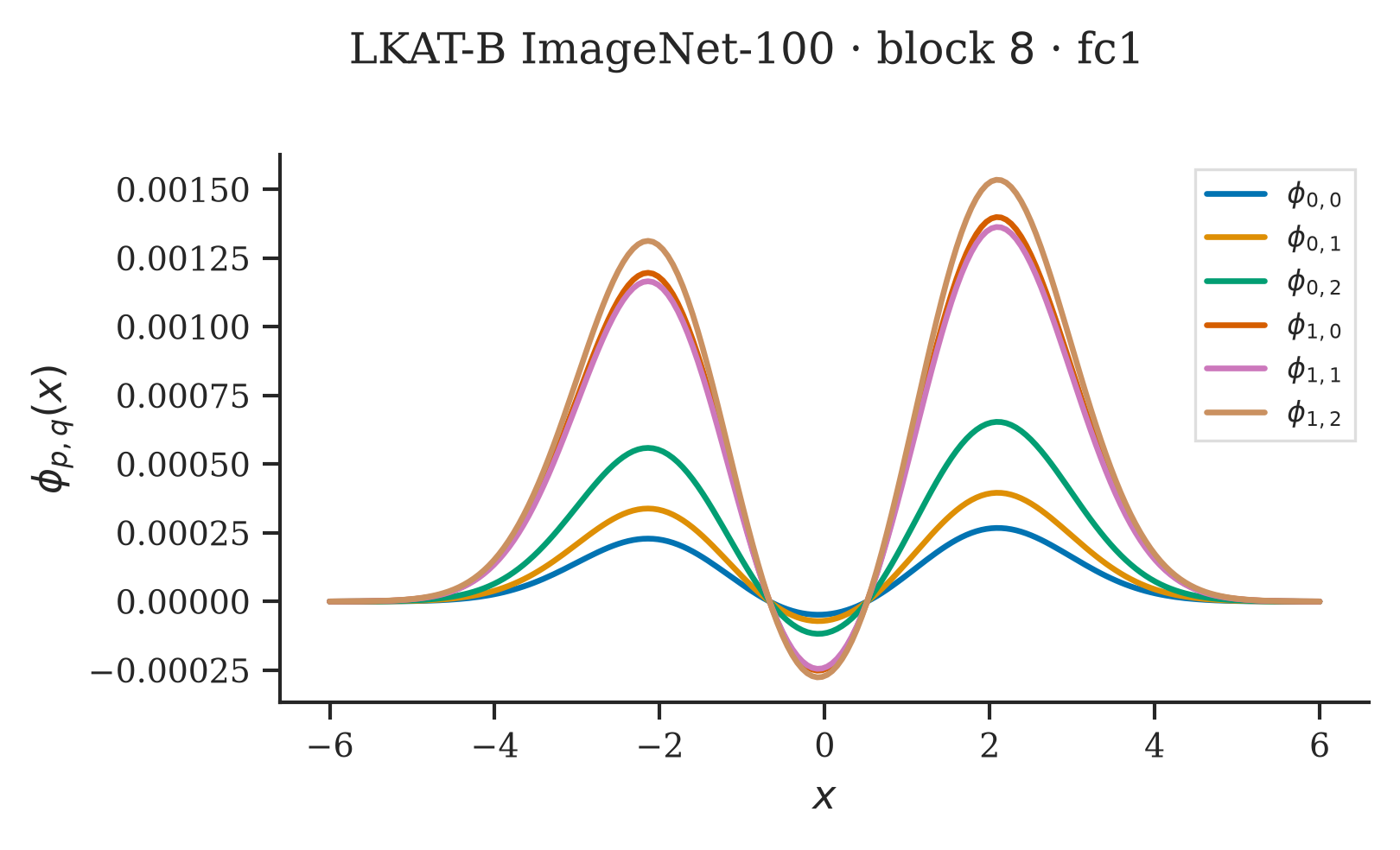}\\[-0.25em]
			{\scriptsize 8}
		\end{minipage}\hspace{0.01\linewidth}
		\begin{minipage}[t]{0.15\linewidth}
			\centering
			\includegraphics[width=\linewidth]{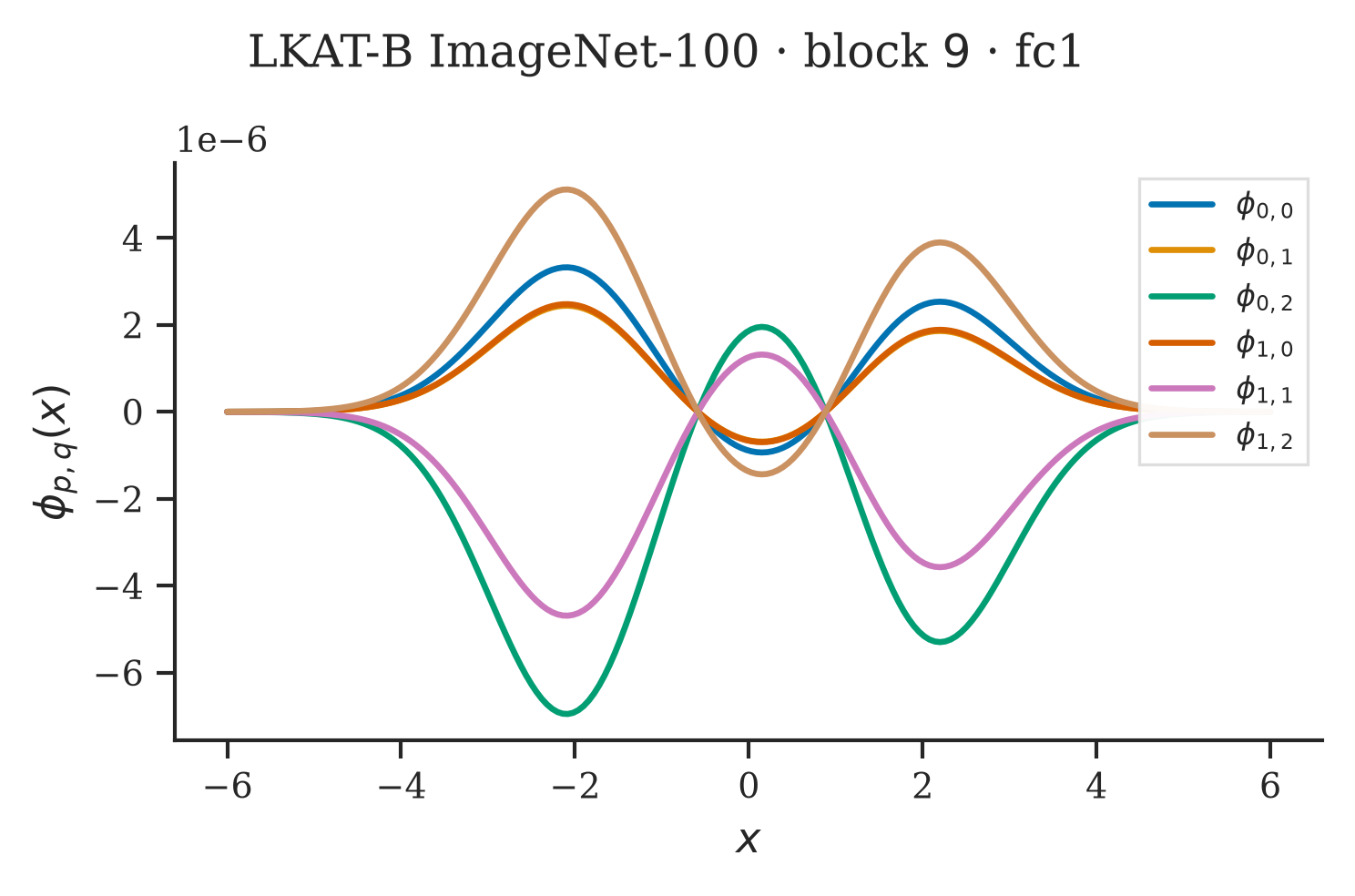}\\[-0.25em]
			{\scriptsize 9}
		\end{minipage}\hspace{0.01\linewidth}
		\begin{minipage}[t]{0.15\linewidth}
			\centering
			\includegraphics[width=\linewidth]{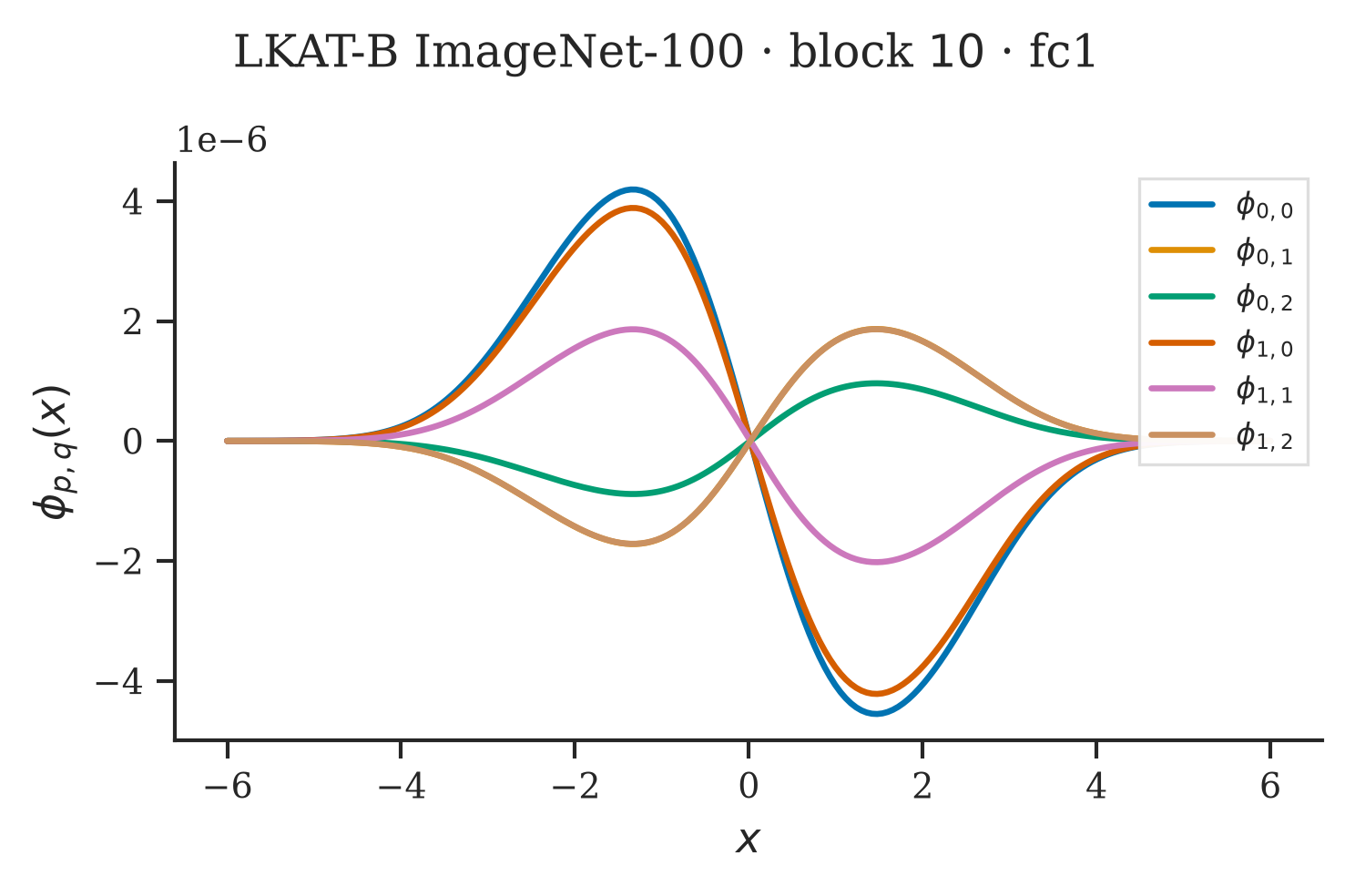}\\[-0.25em]
			{\scriptsize 10}
		\end{minipage}\hspace{0.01\linewidth}
		\begin{minipage}[t]{0.15\linewidth}
			\centering
			\includegraphics[width=\linewidth]{learned_curve/lkat_b_imagenet100/block11_fc1_curves.png}\\[-0.25em]
			{\scriptsize 11}
		\end{minipage}
		\caption{Learned RBF--KAN edge functions $\phi_{p,q}(x)$ for ImageNet-100
		LKAT-B, first KAN layer (fc1), blocks~$0$--$11$ (panel labels).}
		\label{fig:learned-phi-fc1}
	\end{figure}

\begin{figure}[H]
		\centering
		\begin{minipage}[t]{0.15\linewidth}
			\centering
			\includegraphics[width=\linewidth]{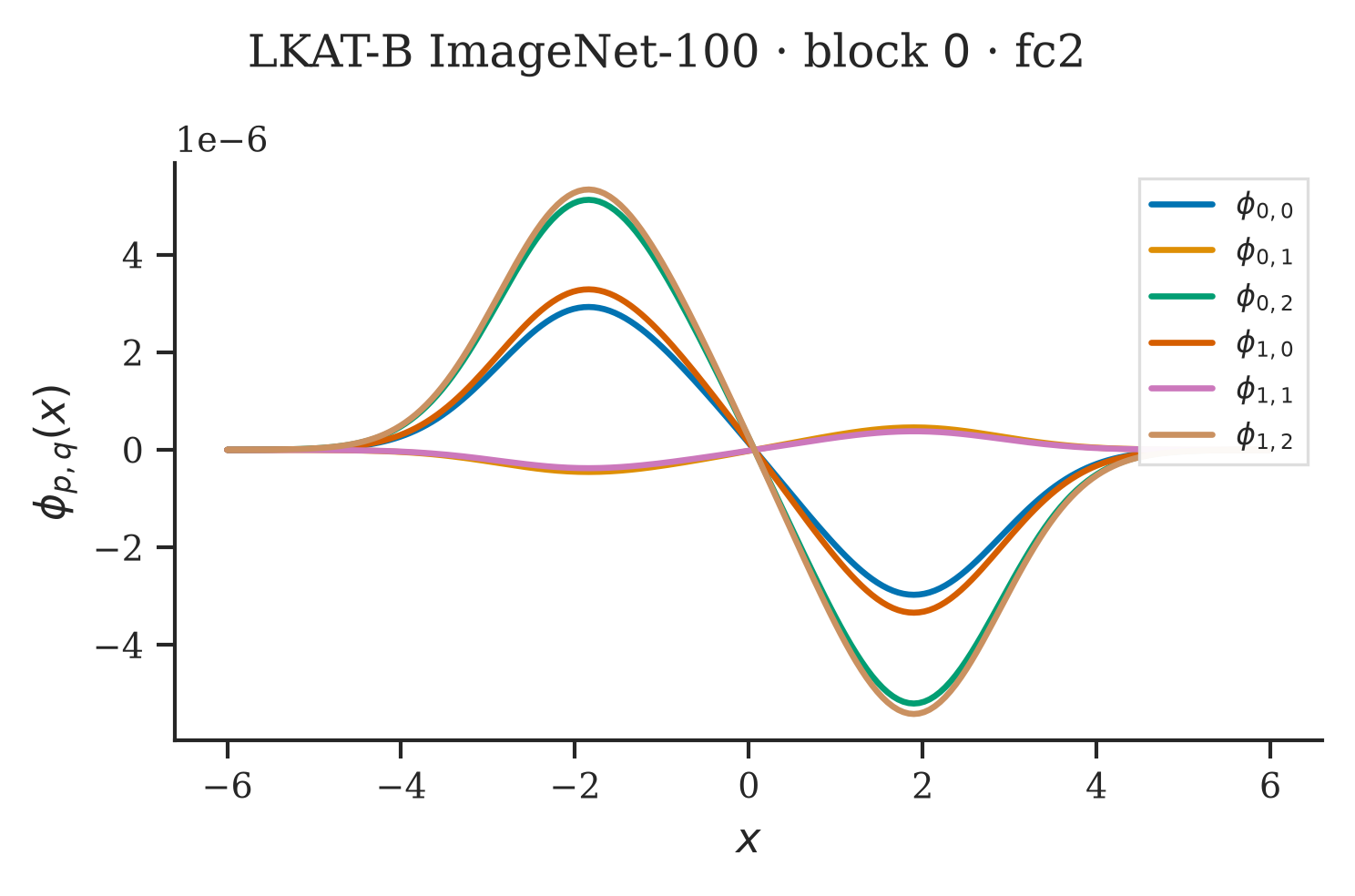}\\[-0.25em]
			{\scriptsize 0}
		\end{minipage}\hspace{0.01\linewidth}
		\begin{minipage}[t]{0.15\linewidth}
			\centering
			\includegraphics[width=\linewidth]{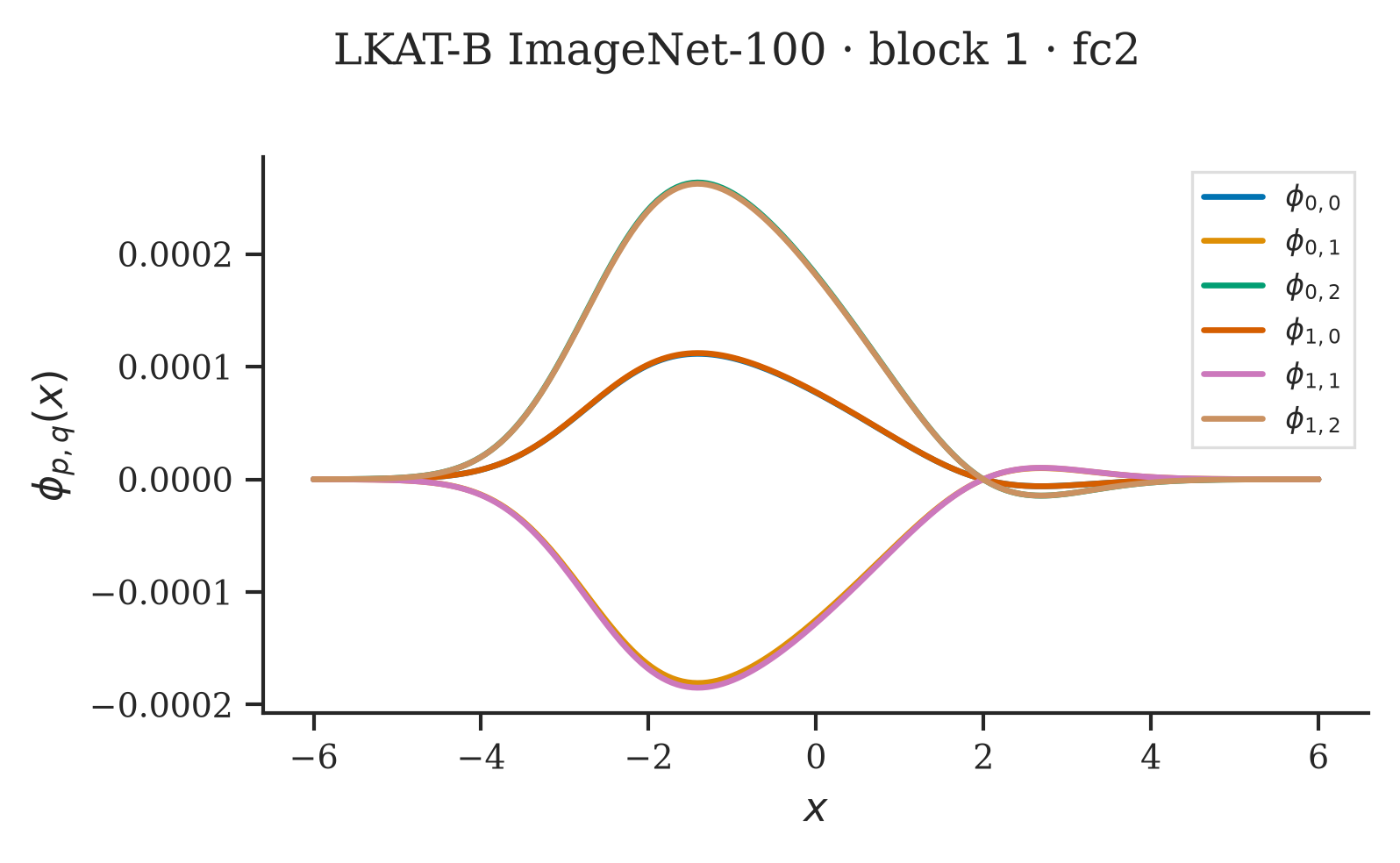}\\[-0.25em]
			{\scriptsize 1}
		\end{minipage}\hspace{0.01\linewidth}
		\begin{minipage}[t]{0.15\linewidth}
			\centering
			\includegraphics[width=\linewidth]{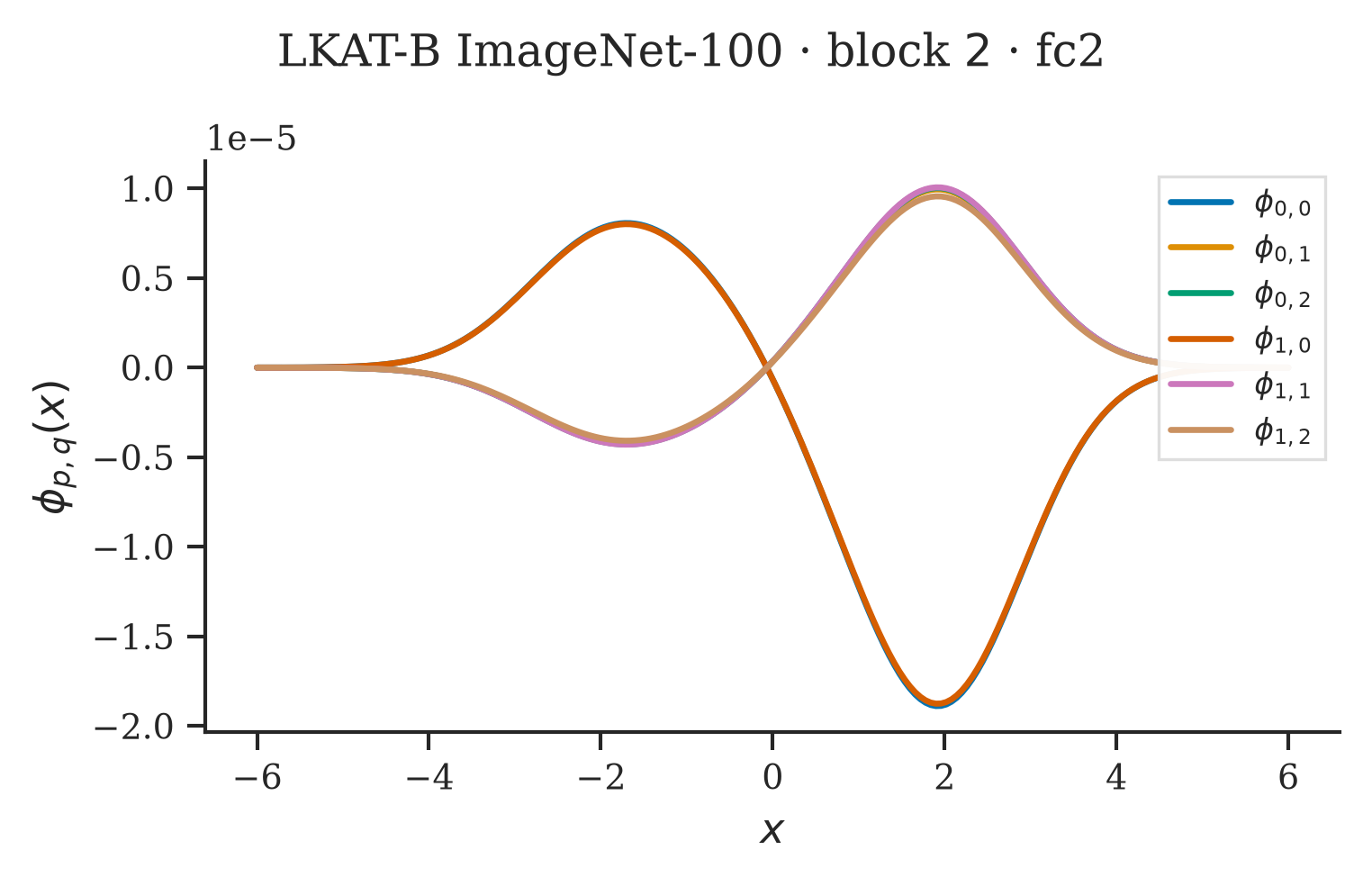}\\[-0.25em]
			{\scriptsize 2}
		\end{minipage}\hspace{0.01\linewidth}
		\begin{minipage}[t]{0.15\linewidth}
			\centering
			\includegraphics[width=\linewidth]{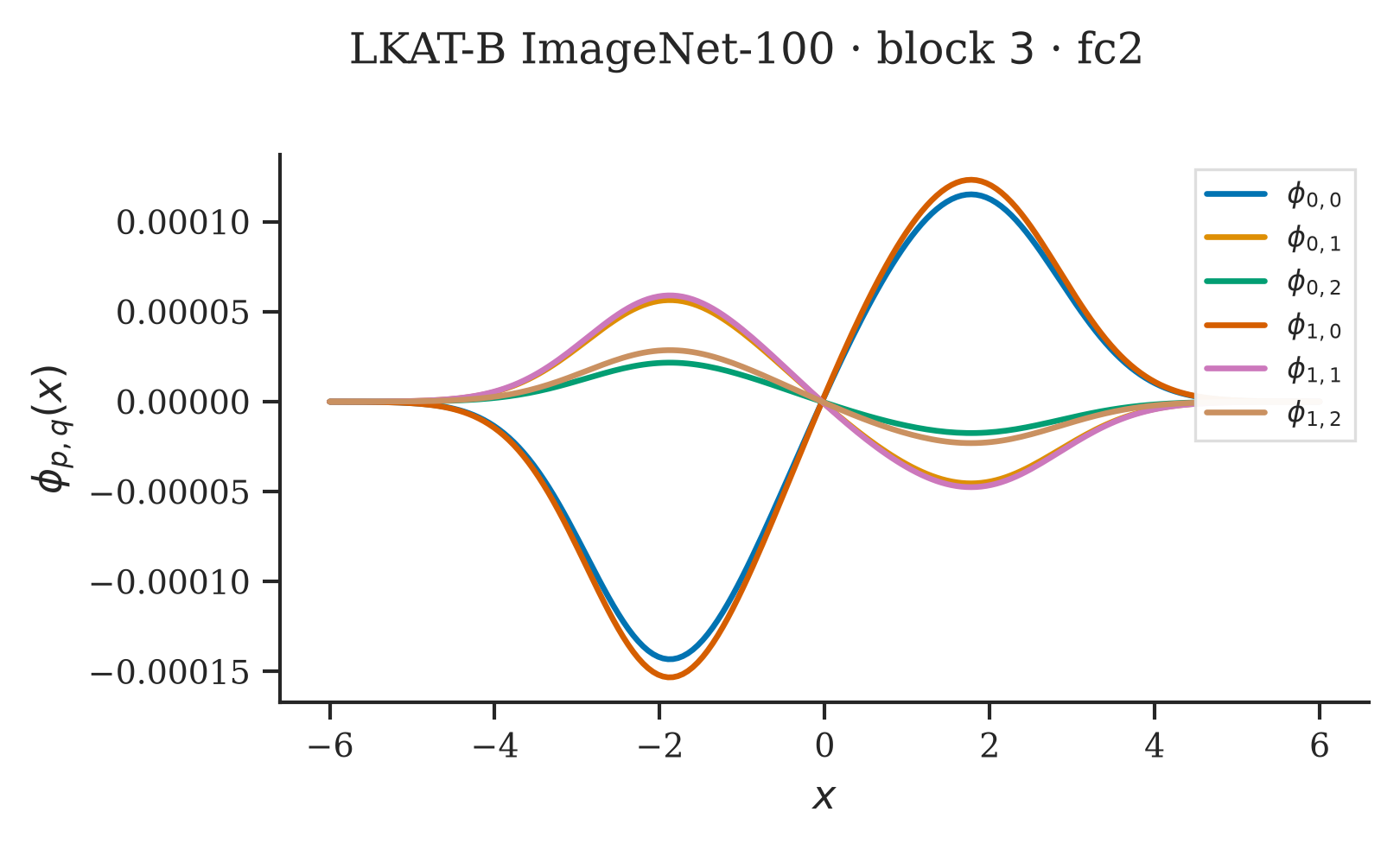}\\[-0.25em]
			{\scriptsize 3}
		\end{minipage}\hspace{0.01\linewidth}
		\begin{minipage}[t]{0.15\linewidth}
			\centering
			\includegraphics[width=\linewidth]{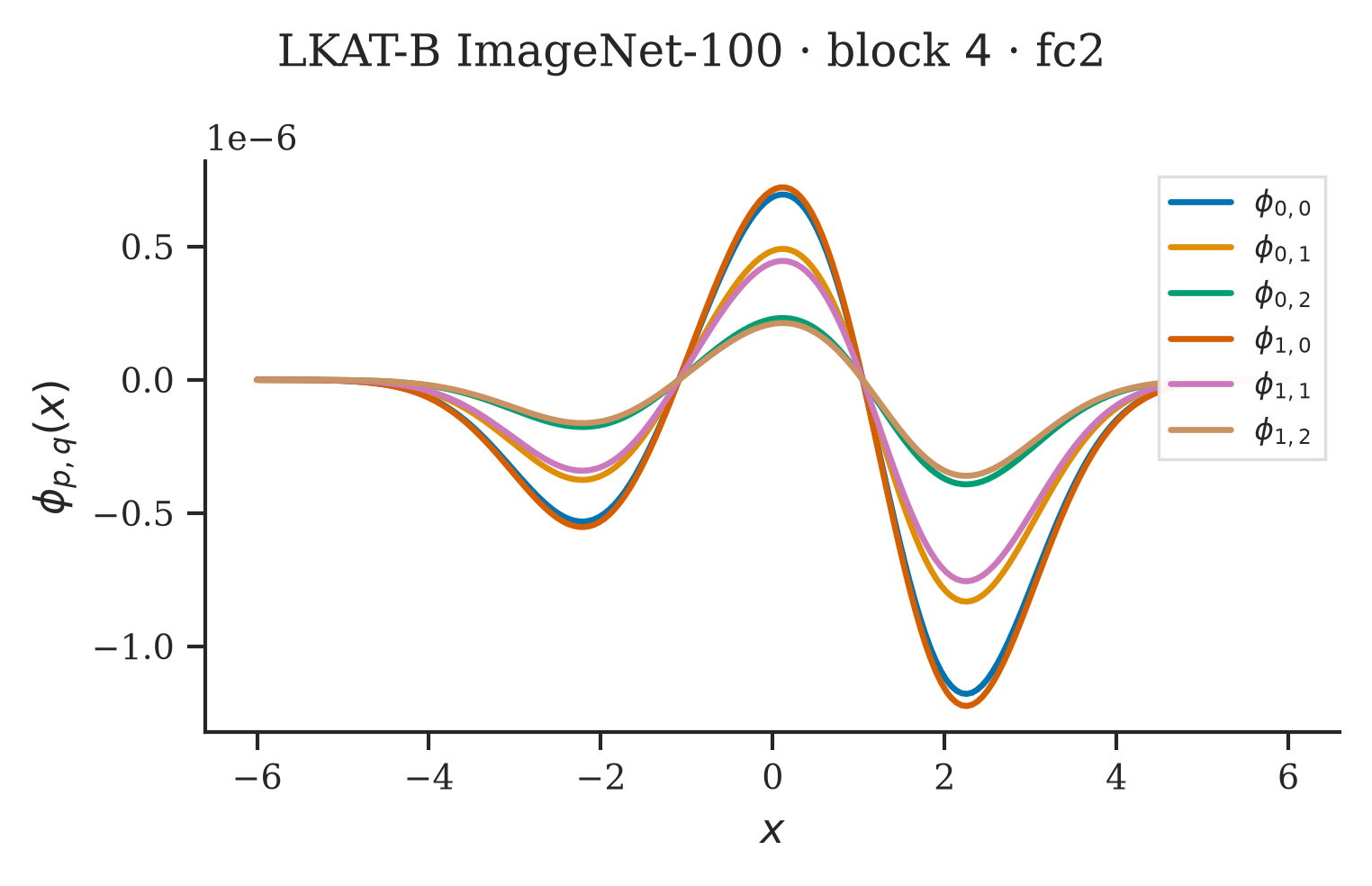}\\[-0.25em]
			{\scriptsize 4}
		\end{minipage}\hspace{0.01\linewidth}
		\begin{minipage}[t]{0.15\linewidth}
			\centering
			\includegraphics[width=\linewidth]{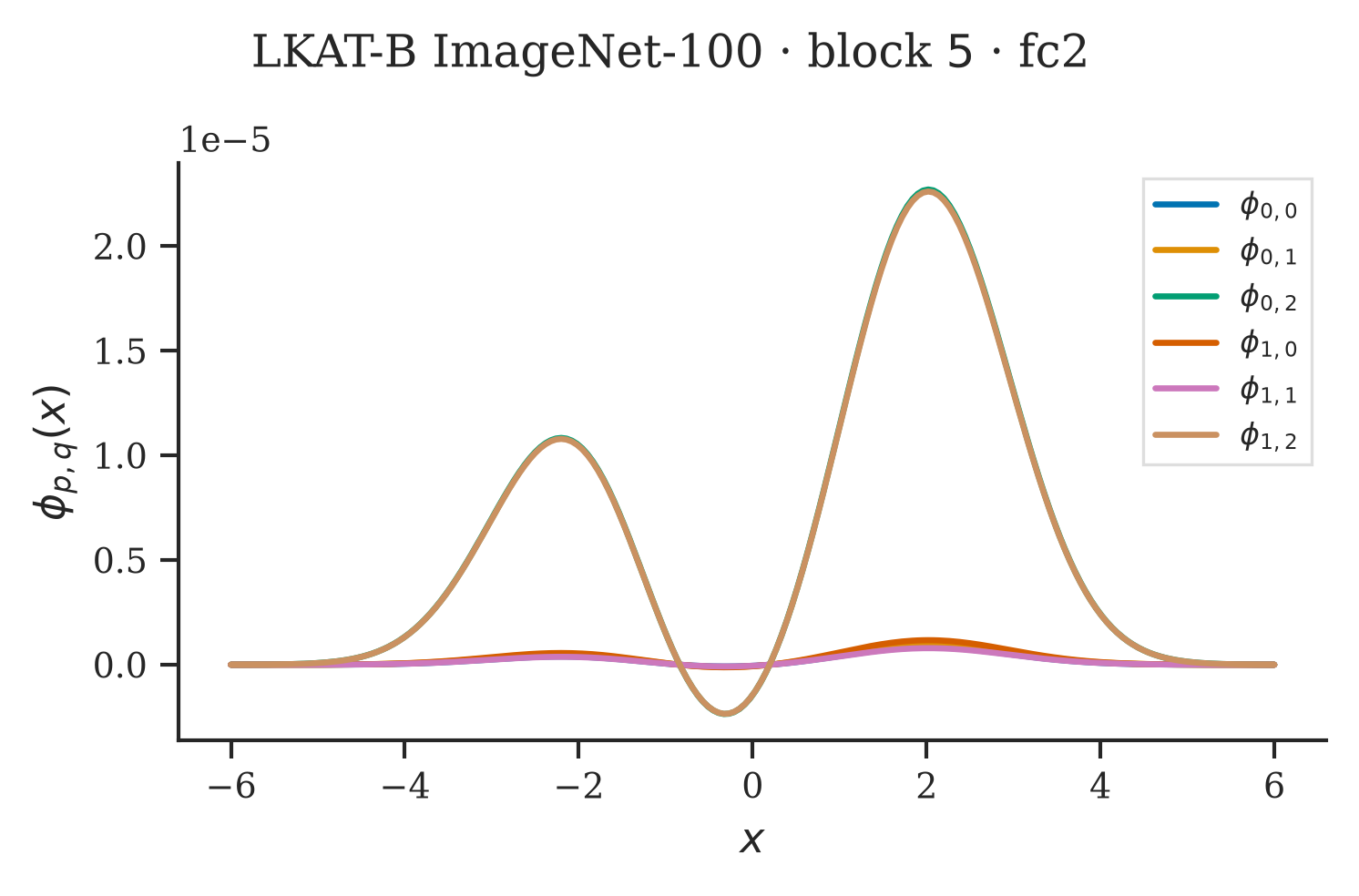}\\[-0.25em]
			{\scriptsize 5}
		\end{minipage}\\[0.35em]
		\begin{minipage}[t]{0.15\linewidth}
			\centering
			\includegraphics[width=\linewidth]{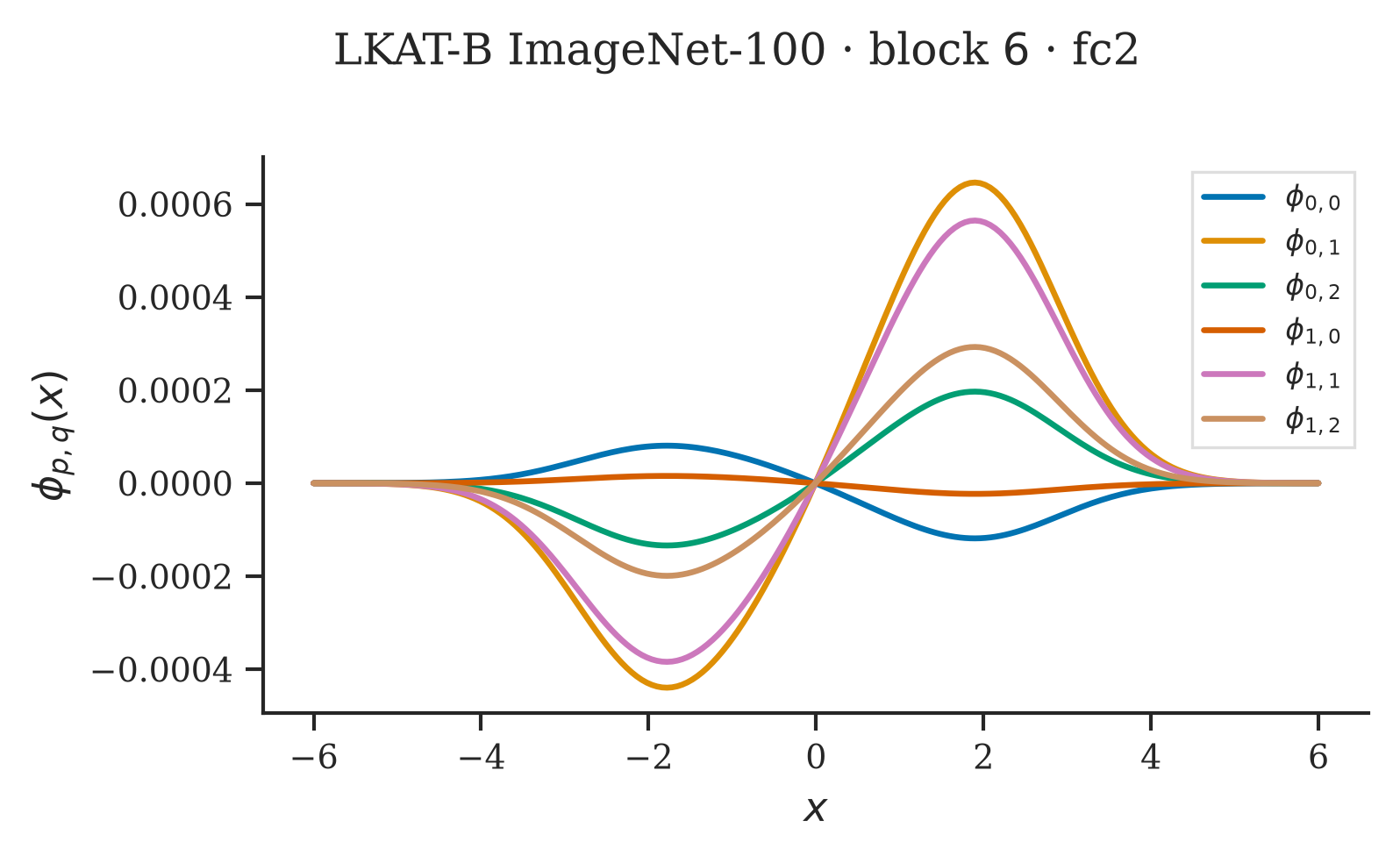}\\[-0.25em]
			{\scriptsize 6}
		\end{minipage}\hspace{0.01\linewidth}
		\begin{minipage}[t]{0.15\linewidth}
			\centering
			\includegraphics[width=\linewidth]{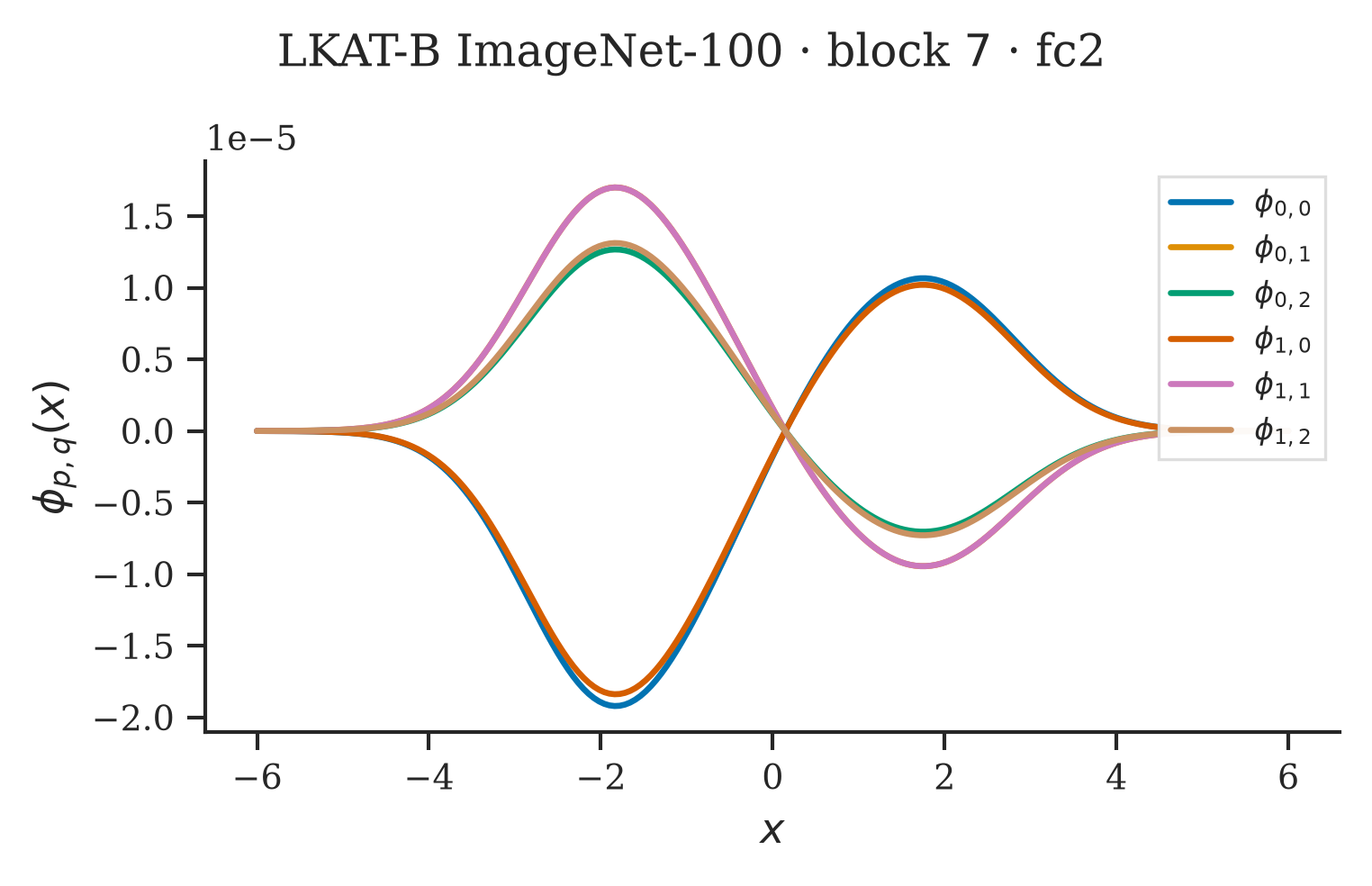}\\[-0.25em]
			{\scriptsize 7}
		\end{minipage}\hspace{0.01\linewidth}
		\begin{minipage}[t]{0.15\linewidth}
			\centering
			\includegraphics[width=\linewidth]{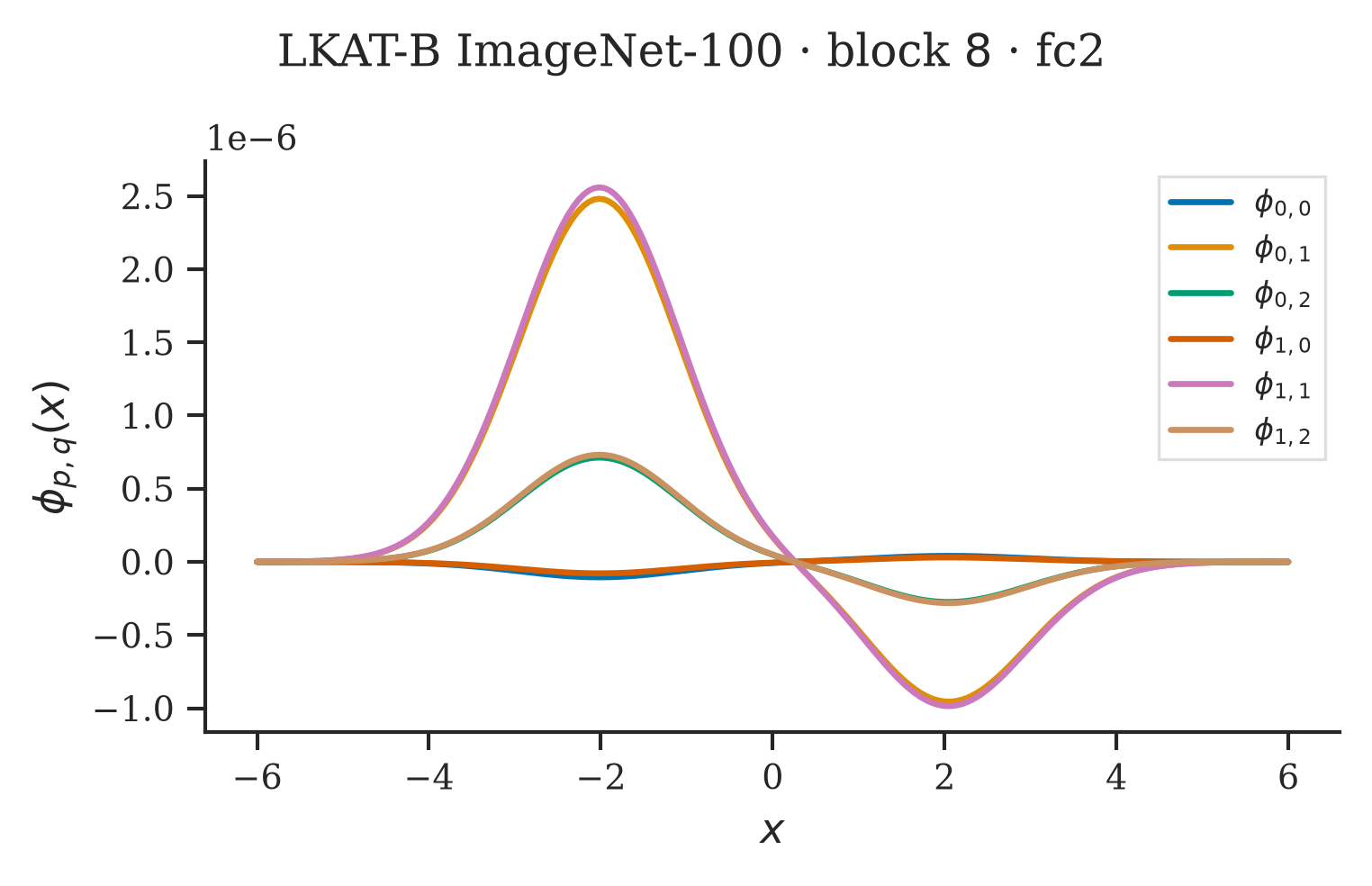}\\[-0.25em]
			{\scriptsize 8}
		\end{minipage}\hspace{0.01\linewidth}
		\begin{minipage}[t]{0.15\linewidth}
			\centering
			\includegraphics[width=\linewidth]{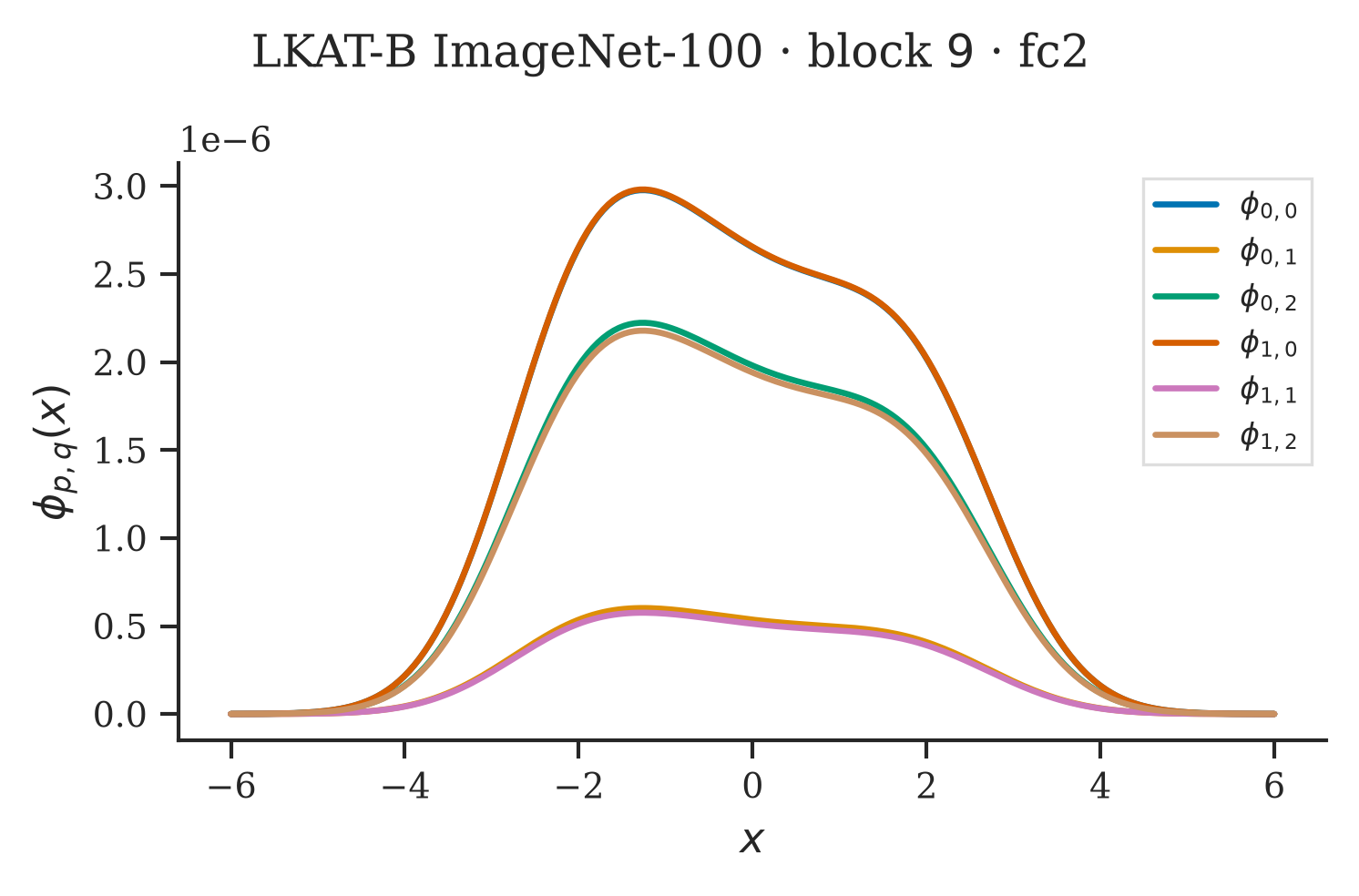}\\[-0.25em]
			{\scriptsize 9}
		\end{minipage}\hspace{0.01\linewidth}
		\begin{minipage}[t]{0.15\linewidth}
			\centering
			\includegraphics[width=\linewidth]{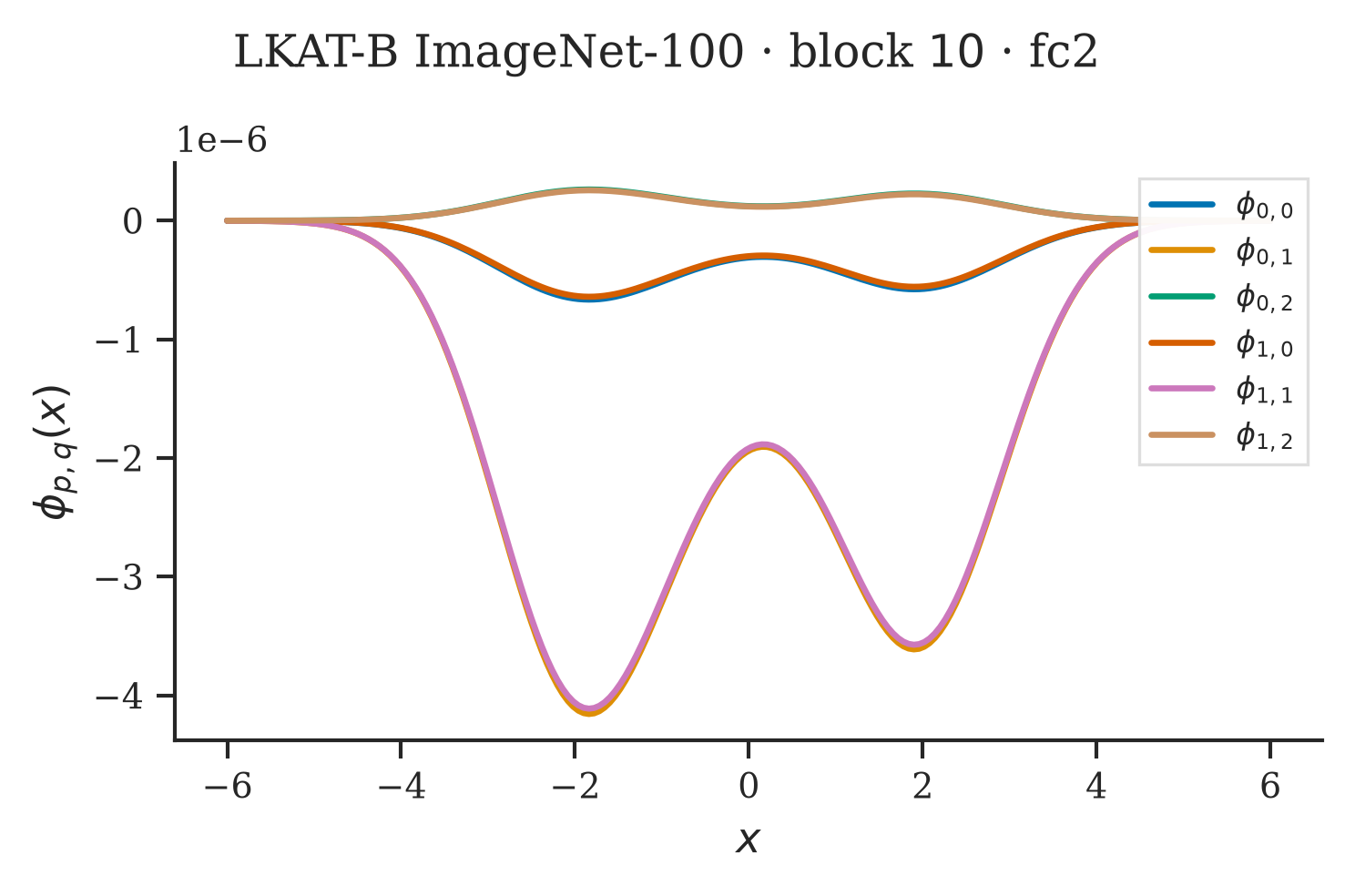}\\[-0.25em]
			{\scriptsize 10}
		\end{minipage}\hspace{0.01\linewidth}
		\begin{minipage}[t]{0.15\linewidth}
			\centering
			\includegraphics[width=\linewidth]{learned_curve/lkat_b_imagenet100/block11_fc2_curves.png}\\[-0.25em]
			{\scriptsize 11}
		\end{minipage}
		\caption{Learned RBF--KAN edge functions $\phi_{p,q}(x)$ for ImageNet-100
		LKAT-B, second KAN layer (fc2), blocks~$0$--$11$ (panel labels).}
		\label{fig:learned-phi-fc2}
	\end{figure}

	Figures~\ref{fig:learned-phi-fc1}--\ref{fig:learned-phi-fc2} extend
	Figure~\ref{fig:learned-phi-main} to all twelve encoder blocks for both
	KAN layers (fc1, fc2) of ImageNet-100 LKAT-B. Across depth, the learned
	univariate functions $\phi_{p,q}$ remain localized on the radial grid and
	continue to change shape, rather than collapsing to a shared fixed
	nonlinearity.
	
	\newpage

	\printbibliography
	
\end{document}